%% file: main.tex
\documentclass{article}

\usepackage{microtype}
\usepackage{graphicx}
\usepackage{subfigure}
\usepackage{booktabs} 

\usepackage{hyperref}

\usepackage[accepted]{icml2024}

\usepackage{amsmath}
\usepackage{amssymb}
\usepackage{mathtools}
\usepackage{amsthm}

\usepackage[capitalize,noabbrev]{cleveref}

\theoremstyle{plain}
\newtheorem{theorem}{Theorem}[section]
\newtheorem{proposition}[theorem]{Proposition}
\newtheorem{lemma}[theorem]{Lemma}
\newtheorem{corollary}[theorem]{Corollary}
\theoremstyle{definition}
\newtheorem{definition}[theorem]{Definition}

\theoremstyle{remark}
\newtheorem{remark}[theorem]{Remark}

\newcommand\normx[1]{\Vert#1\Vert}
\newcommand\ceilx[1]{\lceil#1\rceil}
\newcommand\floorx[1]{\lfloor#1\rfloor}
\DeclareMathAlphabet{\mathcal}{OMS}{cmsy}{m}{n}
\SetMathAlphabet{\mathcal}{bold}{OMS}{cmsy}{b}{n}
\usepackage{dsfont}
\input{math_commands.tex}

\DeclareMathSymbol{\shortminus}{\mathbin}{AMSa}{"39}
\newcommand{\revision}[1]{#1}

\usepackage[textsize=tiny]{todonotes}

\icmltitlerunning{Theoretical Analysis of Learned Database Operations under Distribution Shift through Distribution Learnability}

\begin{document}

\twocolumn[
    \icmltitle{Theoretical Analysis of Learned Database Operations under Distribution Shift through Distribution Learnability}



\icmlsetsymbol{equal}{*}

\begin{icmlauthorlist}
\icmlauthor{Sepanta Zeighami}{ucb}
\icmlauthor{Cyrus Shahabi}{usc}
\end{icmlauthorlist}

\icmlaffiliation{ucb}{University of California, Berkeley. Work done while at USC's Infolab}
\icmlaffiliation{usc}{University of Southern California}

\icmlcorrespondingauthor{Sepanta Zeighami}{zeighami@berkeley.edu}

\icmlkeywords{Machine Learning, ICML}

\vskip 0.3in
]

\setlength{\abovedisplayskip}{5pt}
\setlength{\belowdisplayskip}{5pt}
\setlength{\abovedisplayshortskip}{5pt}
\setlength{\belowdisplayshortskip}{5pt}

\setlength{\textfloatsep}{5pt}
\setlength{\floatsep}{5pt}
\setlength{\intextsep}{5pt}
\setlength{\dbltextfloatsep}{5pt}
\setlength{\dblfloatsep}{5pt}
\setlength{\abovecaptionskip}{5pt}
\setlength{\belowcaptionskip}{5pt}



\printAffiliationsAndNotice{} 

\begin{abstract}
Use of machine learning to perform database operations, such as indexing, cardinality estimation, and sorting, is shown to provide substantial performance benefits. However, when datasets change and data distribution shifts, empirical results also show performance degradation for learned models, possibly to worse than non-learned alternatives. This, together with a lack of theoretical understanding of learned methods undermines their practical applicability, since there are no guarantees on how well the models will perform after deployment. In this paper, we present the first known theoretical characterization of the performance of learned models in dynamic datasets, for the aforementioned operations. Our results show novel theoretical characteristics achievable by learned models and provide bounds on the performance of the models that characterize their advantages over non-learned methods, showing why and when learned models can outperform the alternatives. 
Our analysis develops the \textit{distribution learnability} framework and novel theoretical tools which build the foundation for the analysis of learned database operations in the future.
\end{abstract}

\input{introduction}

\input{preliminaries}

\input{analysis_framework}

\vspace{-0.4cm}
\section{Results}\label{sec:using_oracles}
\input{results}


\input{conclusion}

\section*{Acknowledgements}
This research has been funded by NSF grant IIS-2128661 and NIH grant 5R01LM014026. Opinions,
findings, conclusions, or recommendations expressed in this material are those of the author(s) and
do not necessarily reflect the views of any sponsors, such as NSF and NIH.

\section*{Impact Statement}
\revision{This paper presents work whose goal is to theoretically understand the use of Machine Learning in database systems. Developing such theoretical results can improve trust and usability of ML in database systems, leading to a more widespread adoption in practice. This can help reduce cost, lower energy consumption, and improve usability for database systems. There can be various societal consequences for this, including the creation of more data-driven applications. }

\bibliography{references}
\bibliographystyle{icml2024}

\appendix

\section{Related Work}\label{appx:rel_work}
\input{rel_work}

\section{Formalized Setup and Operations}\label{appx:setup}
\input{setup}

\input{learnability_through_app_formal}

\input{proofs}

\end{document}

%% file: math_commands.tex
\usepackage{amsmath,amsfonts,bm}

\def\eqref#1{equation~\ref{#1}}

\def\1{\bm{1}}

\def\vp{{\bm{p}}}
\def\vq{{\bm{q}}}
\def\vr{{\bm{r}}}

\def\vx{{\bm{x}}}

\def\mD{{\bm{D}}}

\DeclareMathAlphabet{\mathsfit}{\encodingdefault}{\sfdefault}{m}{sl}
\SetMathAlphabet{\mathsfit}{bold}{\encodingdefault}{\sfdefault}{bx}{n}



%% file: introduction.tex
\vspace{-0.8cm}\section{Introduction}\vspace{-0.2cm}
Given a fixed dataset, learned database operations (machine learning models learned to perform database operations such as indexing, cardinality estimation and sorting) have been shown to outperform non-learned methods, providing speed-ups and space savings both empirically \citep{kraska2018case, kipf2018learned, kristo2020case} 
and, for the case of indexing, theoretically \citep{zeighami2023distribution, ferragina2020learned}. For dynamic datasets (e.g., when new points can be inserted into the dataset), significant empirical benefits are also often observed when using learned methods. However, an important caveat accompanying these results is that, especially when data distribution changes, models' performance may deteriorate after new insertions \cite{ding2020alex, negi2023robust, wang2021we}, possibly to worse than non-learned methods~\cite{wongkham2022updatable}. This, combined with the lack of a theoretical understanding of the behavior of the learned models as datasets change, poses a critical hurdle to their deployment in practice. It is theoretically unclear why and when learned models outperform non-learned methods, and, until this paper, no theoretical work shows any advantage in using the learned methods in dynamic datasets and under distribution shift. The goal of this paper is to theoretically understand the capabilities of learned models for database operations, show why and when they outperform non-learned alternatives and provide theoretical guarantees on their performance. 

We specifically study learned solutions for three fundamental database operations: indexing, cardinality estimation and sorting. Our main focus is the study of learned indexing and cardinality estimation in the presence of insertions from a possibly changing data distribution, while we also study learned sorting (in static scenario) to show the broader applicability of our developed theoretical tools. 
In all cases, a learned model, $\hat{f}(x;\theta)$ is used to replace a specific data operation, $f_{\mD}(x)$, that takes an input $x$ and calculates a desired answer from the dataset $\mD$. For cardinality estimation, $f_{\mD}(x)$ returns the number of points in the database $\mD$ that match the query $x$, and for indexing $f_{\mD}(x)$ returns the true location of $x$ in a sorted array. The model $\hat{f}(x;\theta)$ is trained to approximate $f_{\mD}(x)$, and an accurate approximation 
leads to efficiency gains when using the model (e.g., for learned indexing, if $\hat{f}(x;\theta)$ gives an accurate estimate of location of $x$ in a sorted array, a local search around the estimated location efficiently finds the exact location). In the presence of insertions, the ground-truth $f_{\mD}(x)$ changes as the dataset changes (e.g., the cardinality of some queries increase as new points are inserted). Thus, as more points are inserted (not only due to distribution shift, but exacerbated by it), the accuracy of $\hat{f}(x;\theta)$ worsens. A common solution is to periodically retrain $\hat{f}$ to ensure consistent accuracy. This, however, increases insertion cost when insertions trigger a (computationally expensive) model retraining. 

\begin{table*}[t!]
    \centering
    \vspace{-0.1cm}
    \begin{tabular}{l l l l}
    \toprule
       \textbf{Learned Operation}  & \textbf{Query Complexity} & \textbf{Insertion Complexity} & \textbf{Space Complexity} \\\midrule
       \textbf{Indexing}  & $\mathcal{T}_n^{\mathfrak{X}}\log\log n+\log\delta n$  & $\mathcal{T}_n^{\mathfrak{X}}\log\log n+\log\delta n+\mathcal{B}_n^\mathfrak{X}\log^2\log n$ & $n\log n^\dagger$\\
       \textbf{CE}, $d$-dim, $\epsilon=\Omega(\sqrt{n})$  & $\mathcal{T}_n^{\mathfrak{X}}$ & $\max\{\frac{\delta n}{\epsilon}, 1\}\mathcal{B}_n^{\mathfrak{X}}$ & $\mathcal{S}_n^{\mathfrak{X}}$ \\
       \textbf{CE}, $1$-dim  & $\mathcal{T}_{\epsilon^2}^\mathfrak{X}+\log n$ & $\max\{\delta\epsilon, 1\}\mathcal{B}_{\epsilon^2}^\mathfrak{X}+\log n$ & $\frac{n}{\epsilon^2}\mathcal{S}_{\epsilon^2}^\mathfrak{X}+\frac{n}{\epsilon^2}\log n$ \\\midrule
       \textbf{Sorting}  & \multicolumn{2}{l}{$\mathcal{T}_{n}^\mathfrak{X}n\log\log n^\dagger$}  & $\mathcal{S}_{\sqrt{n}}^\mathfrak{X}+n\log n$ \\
       \textbf{Sorting}, appx. known dist.  & \multicolumn{2}{l}{$\mathcal{T}^\mathfrak{X}n$}  & $\mathcal{S}^\mathfrak{X}$ \\
       \bottomrule
    \end{tabular}
    \vspace{-0.2cm}
    \caption{Summary of results for data sampled from a distribution learnable class $\mathfrak{X}$ (CE: cardinality estimation, $\dagger$: for simplicity assuming $\mathcal{S}_n^{\mathfrak{X}},\mathcal{B}_n^{\mathfrak{X}}$ are at most linear in data size, see Theorem~\ref{thm:dynamic_indexing_oracle} and Theorem~\ref{thm:sorting_oracle} for general cases).}
    \label{tab:res_sum}
\end{table*}

Theoretically, the relationship between accuracy change and new data insertion has not been well understood, leading to a lack of meaningful theoretical guarantees for learned methods in the presence of insertions. 
The only existing guarantees are by the PGM index~\citep{ferragina2020pgm}, which achieves a worst-case insertion time of $O(\log n)$ with worst-case query time of $O(\log^2 n)$. Although experimental results show PGM often outperforms B-trees in practice \cite{ferragina2020pgm, wongkham2022updatable}, the theoretical guarantees are worse than those of a B-tree (that supports both insertions and queries in $O(\log n)$). Such theoretical guarantees do not meaningfully characterize the index's performance in practice nor show why and when the learned model performs better (or worse) than B-trees. 

In this paper, we present the first known theoretical characterization of the performance of learned models for indexing and cardinality estimation in the presence of insertions, painting a thorough picture of why and when they outperform non-learned alternatives for these fundamental database operations. 
Our analysis develops the notion of \textit{distribution learnability}, a characteristic of data distributions that helps quantify learned database operation's performance for data form such distributions. Using this notion, our results are distribution dependent (as one expects bounds on learned operations should be), without making unnecessary assumptions about data distribution. Our developed theoretical framework builds a foundation for the analysis of learned database operations in the future. To show its broader applicability, we present a theoretical analysis of learned sorting, showing its theoretical characteristics and proving why and when it outperforms non-learned methods.


\vspace{-0.25cm}
\subsection{Summary of Results}\label{sec:summary_res}
\vspace{-0.1cm}
Table~\ref{tab:res_sum} summarizes our results in the following setting. Suppose $n$ data points are sampled independently from distributions $\chi_1, ..., \chi_n$, and let the distribution class $\mathfrak{X}=\{\chi_1, ..., \chi_n\}$. The points are inserted one by one into a dataset. $\mD^i$ denotes the dataset after $i$ insertion. Our goal is to efficiently answer cardinality estimation and indexing queries on $\mD^i$ accurately for any $i$, i.e., as new points are being inserted. We denote \textit{distribution shift} by $\delta\in[0, 1]$ (defined based on, and often equal, to total variation distance) where $\delta=0$ means no distribution shift. Table~\ref{tab:res_sum} also contains results for sorting, where the goal is to sort the fixed array $\mD^n$, and the reported results are the time and space complexity of doing so. For sorting only, we assume the samples are i.i.d. All results are expected complexities, with the expectation over sampling of the data, and the insertion complexity is amortized over $n$ insertions. 
To obtain our results, we develop a novel theoretical framework,  dubbed {\it distribution learnability}. We provide an informal discussion of the framework before discussing the results.

\textbf{Distribution Learnability}. At a high level, distribution learnability means \textit{we can model a data distribution well}. This notion allows us to state our results in the form ``\textit{if we can model a data distribution well, learned database operations will perform well on data from that distribution}''. Then, if one indeed proves that ``\textit{we can model the data distribution $\chi$ well}'', our result immediately implies ``\textit{learned database operations will perform well on data coming from $\chi$}''. Crucially, our Theorem~\ref{thm:generic_oracle_construction} shows that purely function approximation results (independent of the application of learned databases) imply distribution learnability, enabling us to utilize function approximation results to show the benefits of learned database operations. 

More concretely (but still informally), we say a distribution class $\mathfrak{X}$ is \textit{distribution learnable} with parameters $\mathcal{T}_n^{\mathfrak{X}}$, $\mathcal{S}_n^{\mathfrak{X}}$, $\mathcal{B}_n^{\mathfrak{X}}$, if given a set of observations, $\mD^n$, from distributions in $\mathfrak{X}$, there exists a learning algorithm that returns an \textit{accurate} model, $\hat{f}$, of the distributions in $\mathfrak{X}$, and that $\hat{f}$ can be evaluated in $\mathcal{T}_n^{\mathfrak{X}}$ operations, and takes space at most $\mathcal{S}_n^{\mathfrak{X}}$ to store. Furthermore, the learning algorithm takes time $n\times\mathcal{B}_n^{\mathfrak{X}}$ to learn $\hat{f}$, where $\mathcal{B}_n^{\mathfrak{X}}$ is the amortized training time. The notion is related to statistical estimation, but we also utilize it to characterize time and space complexity of modeling.  

Results in Table~\ref{tab:res_sum} are stated for data sampled from any distribution learnable class $\mathfrak{X}$. For illustration, we summarize the results for two specific distribution classes: (1) distributions, $\mathfrak{X}_{\rho}$, with p.d.f bounded between $0<\rho_1$ and $\rho_2<\infty$, and (2) distributions, $\mathfrak{X}_{c}$, where the data distribution is known and probability of events can be calculated efficiently (e.g., distribution is known to be uniform or piece-wise polynomial). The first case formulates a realistic scenario for the data distribution (experimentally shown by ~\citet{zeighami2023distribution}), while the second case presents a best case scenario for the learned models, showing what is possible in favorable circumstances. Lemma~\ref{lemma:specific_oracle_construction} shows $\mathfrak{X}_{\rho}$, (and trivially) $\mathfrak{X}_{c}$ are distribution learnable, deriving the corresponding values for $\mathcal{T}_n^{\mathfrak{X}}$, $\mathcal{S}_n^{\mathfrak{X}}$ and $\mathcal{B}_n^{\mathfrak{X}}$ (See Table~\ref{tab:specific_dist_classes} for exact values).
Next,  we discuss Table~\ref{tab:res_sum} for $\mathfrak{X}_{\rho}$ and $\mathfrak{X}_{c}$, where we substitute the values of $\mathcal{T}^{\mathfrak{X}}_n$, $\mathcal{S}^{\mathfrak{X}}_n$ and $\mathcal{B}^{\mathfrak{X}}_n$ from Lemma~\ref{lemma:specific_oracle_construction} for $\mathfrak{X}_\rho$ and $\mathfrak{X}_c$, and discuss the resulting complexities. 




\textbf{Indexing}. After substituting the complexities in the first row of Table~\ref{tab:res_sum} we obtain that for $\mathfrak{X}_\rho$ and $\mathfrak{X}_c$, query and insertion complexities are $O(\log\log n+\log(\delta n))$. To understand this result, consider the simple scenario with $\delta=0$, where inserted items are sampled form a fixed distribution, and thus frequent model updates are not necessary. The result shows that a learned model performs insertions and queries in $O(\log\log n)$, showing their superiority over non-learned methods that perform queries and insertions in $O(\log n)$. Nonetheless, when there is a distribution shift, model performance worsens. In the worst-case and when $\delta=1$, we see no advantage to using learned models over non-learned methods. This is not surprising, since learned models use current observations to make prediction about the future, and if the future distribution is drastically different, one should not be able to gain from using the current observations. 

\textbf{Cardinality Estimation}. First, consider the second row in Table.~\ref{tab:res_sum}, showing performance of learned models for cardinality estimation in high dimensions but when error is at least $\sqrt{n}$. Substituting the complexities in this row, for $\mathfrak{X}_c$, we obtain that learned models perform insertions and queries in $O(1)$ time and space in this setting. This is significant, given that a non-learned method such as sampling (and more broadly $\epsilon$-approximations~\citep{mustafa2017epsilon}), even in this accuracy regime, needs space exponential in dimensionality \citep{wei2018tight}. Nonetheless, modeling in high dimensions is difficult, and consequently this result requires the accuracy to be at least $\sqrt{n}$. Moreover, even for $\epsilon\geq\sqrt{n}$ but for more general distribution class of $\mathfrak{X}_\rho$, our results show that learned methods will also take space exponential in dimensionality (which is broadly needed, even for neural networks \citep{petersen2018optimal}, without further assumptions). 
We also mention that $\sqrt{n}$ has a statistical significance (see Sec.~\ref{sec:analysis_framework} for discussion), and appears in our analysis throughout. 
Second, we show that in 1-dimension (the third row of Table~\ref{tab:res_sum}), learned models perform cardinality estimation queries effectively, where for $\mathfrak{X}_c$, a learned model can perform queries and insertions in $O(\log n)$ while taking space $O(\frac{n}{\epsilon^2}\log n)$. This result also shows that a learned approach outperforms the non-learned (and worst-case optimal) method discussed in \citep{wei2018tight} that takes space $O(\frac{n}{\epsilon}\log \epsilon n)$ to answer queries. 

\textbf{Sorting}. 
Substituting complexities in the fourth row of Table~\ref{tab:res_sum}, we obtain $O(n\log \log n)$ time complexity for $\mathfrak{X}_\rho$, using a method that is a variation of ~\citet{kristo2020case} that learns to sort through sampling. Our framework applies to this method because its study needs to consider the generalization of a model learned from samples (similar to how models need to generalize to a new dataset after insertions). Moreover, last row of Table~\ref{tab:res_sum} shows that, if we (approximately) know the data distribution, and the distribution can be efficiently evaluated and stored, we can sort an array in $O(n)$ ($\mathcal{T}^\mathfrak{X}$ is independent of data size), showing benefits of using data distribution to perform database operations.  

To conclude, our results in Table~\ref{tab:res_sum} are more general than the two distribution classes discussed above. A major contribution of this paper is developing the distribution learnability framework that allows us to orthogonally study the two problems of modeling a data distribution (\textit{the modeling problem}), and how learned models can be used to perform database operations with theoretical guarantees (\textit{the model utilization problem}). Table~\ref{tab:res_sum} summarizes our contributions to the latter problem, while our results connecting distribution learnability to function approximation concepts (Theorem~\ref{thm:generic_oracle_construction}) is our contribution to the former. The rest of this paper discusses our developed framework and results in more detail, but for the sake of space, formal discussion, proofs, and a detailed discussion of the related work are differed to the appendix.

%% file: preliminaries.tex
\vspace{-0.4cm}
\section{Preliminaries}
\vspace{-0.1cm}
\subsection{Problem Setting}
\vspace{-0.1cm}
\textbf{Setup}. We study performing database operations on a possibly changing $d$-dimensional dataset. We either consider the setting when $n$ data points are inserted one by one (dynamic setting), or that we are given a fixed set of $n$ data points (static setting). We define $\mD^i$ as the dataset after $i$ insertions, and the final dataset, $\mD^n$, is often denoted as $\mD$. We study indexing, cardinality estimation and sorting operations.  

For indexing, the goal is to build an index to store and find items in a $1$-dimensional dataset. The index supports insertions and queries. That is, after $i$ insertions, for any $i$, we can retrieve items from the dataset $\mD^i$, where the query is either an exact match query or a range query. For cardinality estimation, the dataset is $d$-dimensional, and we support insertions and axis-parallel queries. That is, after $i$ insertions, for any $i$, we would like to estimate the number of items in the dataset $\mD^i$ that match a query $q$, where $q$ defines an axis-parallel hyper-rectangle. Finally, the goal of sorting is to sort a fixed 1-dimensional array, $\mD$, of size $n$. Indexing and sorting always return exact results (i.e., array has to be fully sorted after the operation), while cardinality estimation accepts an error of $\epsilon$ for the query answer estimates.

\textbf{Data Distribution and Distribution Shift}. We consider the case that the $i$-th data point is sampled independently from a distribution $\chi_i$, and denote by $\mD\sim\chi$ this sampling procedure, where $\chi=\{\chi_1, ..., \chi_n\}$. We say $\mD$ was sampled from a distribution class $\mathfrak{X}$ if $\chi_i\in\mathfrak{X}$ $\forall i$. We use total variation to quantify distribution shift. We say $\mD$ was sampled from a distribution $\chi$ with distribution shift $\delta$, when $\max_{
\chi_i, \chi_j\in\chi}\normx{\chi_i-\chi_j}_{TV}$, where $\normx{\chi_i-\chi_j}_{TV}$ denoted the total variation (TV) distance between $\chi_i$ and $\chi_j$. We also define total variation of a distribution set $\chi$ as $TV(\chi)=\sup_{
\chi_i, \chi_j\in\chi}\normx{\chi_i-\chi_j}_{TV}$. TV is a number between 0 and 1 with $\delta=1$ the maximum distribution shift and $\delta=0$ the case with no distribution shift.

\textbf{Problem Definition}. We study the performance of learned models when performing the above data operations. Assume an algorithm takes at most  $T_I(\mD)$ operations to perform $n$ insertions from a dataset $\mD$, at most $T_{Q}(\mD)$ to perform any query and has $S(\mD)$, space overhead (excluding the space to store the data).  We study amortized expected insertion time defined as $\frac{1}{n}\mathds{E}_{\mD\sim\chi}T_I(\mD)$, expected query time, ${\mathds{E}_{\mD\sim\chi}}{T_Q(\mD)}$, and storage space, ${\mathds{E}_{\mD\sim\chi}}{S(\mD)}$.

\vspace{-0.2cm}
\subsection{Learned Database Operations}
\vspace{-0.1cm}
\textbf{Operation Functions}. Let $f_\mD(x)$ be an \textit{operation function}, defined as a function that takes an input $x$ and outputs the answer, calculated from the database $\mD$, for some desired operation. In this paper,  $f_{\mD}$ is either the cardinality function, $c_{\mD}(x)$, that takes a query, $x$, as input and outputs the number of points in $\mD$ that match $x$, or the rank function, $r_{\mD}(x)$ that takes a 1-dimensional query as input and returns the number of elements in $\mD$ smaller than $x$.  
The rank function is used in sorting and indexing, because $r_{\mD}(x)$ is the index of $x$ if $\mD$ was stored in a sorted array. We use the notation $f_{\mD}\in \{r_{\mD}, c_{\mD}\}$ (or $f\in\{r, c\}$) to refer to both functions, $r_{\mD}$ and $c_{\mD}$ (for instance, $f_D\geq 0$ is equivalent to the two independent statements that $r_{\mD}\geq 0$ and $c_{\mD}\geq 0$). 

We also define \textit{distribution operation function}, $f_\chi$, for an operation $f$, defined as $f_\chi(\vx)=\mathds{E}_{\mD\sim\chi}[f_\mD(\vx)]$, if $\mD$ is sampled from a distribution $\chi$. Note that \textit{distribution operation function} depend only on the data distribution (and not observed dataset). For instance, $\frac{1}{n}r_\chi$ is the c.d.f of data distribution, $\chi$ if $\mD$ is sampled i.i.d from $\chi$, and similarly $\frac{1}{n}c_\chi(\vx)$ is the probability that a sample from $\chi$ falls in an axis-parallel rectangle defined by $\vx$. We call $\mathds{E}_{\mD\sim\chi}[c_\mD(\vx)]$ distribution cardinatliy function.

\textbf{Learned Database Operations with Insertions}. Learned database operations learn a model $\hat{f}$ that approximates $f_\mD$ well, and use the learned model to obtain an estimate of the operation output (for sorting and indexing, a refinement step ensures exact result, through either local binary search or lightweight sorting). However, as new data points are inserted and the dataset changes, the ground truth answers to operations change, thereby increasing the model error. Note that model answers are scaled to current data size (i.e., if $\hat{f}$ was trained on a dataset of size $i$, and tested on a dataset of size $j$, we report $\frac{j}{i}\hat{f}$ as answers), but this does not stop the error from increasing. Thus, to guarantee the error is below a threshold, one needs to update the models as the datasets change, which is often done by periodically retraining the models. Model retraining contributes to insertion cost in the database. To keep the insertion cost low, one needs to minimize retraining frequency. Meanwhile, infrequent retraining increases error (and, for indexing, query time). Finding a suitable balance between insertion time, accuracy and query time is a subject in much of our theoretical study.

%% file: analysis_framework.tex
\vspace{-0.3cm}
\section{Analysis through Distribution Learnability}\label{sec:analysis_framework}
\vspace{-0.2cm}
Our goal is to ensure that a model $\hat{f}$ trained to perform operations $f$, $f\in\{r, c\}$, has bounded error. We first discuss a lower bound on the error of models in the presence of insertions, which motivates our analysis framework. 


\textbf{Lower Bound on Model Generalization.} Consider a model $\hat{f}$, trained after the $i$-th insertion and using dataset $\mD^i$. Assume the model is not retrained after $k$ further insertions so that $\hat{f}$ is used to answer queries for dataset $\mD^j$, $j=i+k$. The following lemma shows a lower bound on the expected maximum \textit{generalization error} of the model to dataset $\mD^j$, defined as  $\sup_\vx\mathds{E}_{\mD\sim\chi}[|\frac{j}{i}\hat{f}(\vx)-f_{\mD^j}(\vx)]$.


\begin{theorem}\label{lemma:sqrt_n_lower_bound}
    Consider any model $\hat{f}$ trained after the $i$-th insertion and on dataset $\mD^i$. For any integer $j>i+2$ and after performing $k=j-i$ new insertions we have
    $$\sup_\vx\mathds{E}_{\mD^j\sim\chi}[|\frac{j}{i}\hat{f}(\vx)-f_{\mD^j}(\vx)|]\geq \frac{\sqrt{k}}{4},$$
    when $\mD^j$ is i.i.d from any continuous distribution $\chi$. 
\end{theorem}\vspace{-0.3cm}
Theorem~\ref{lemma:sqrt_n_lower_bound} states that the expected error of a \textit{single fixed} model, no matter how good the model is when it is trained, after $k$ insertions, will increase to $\Omega(\sqrt{k})$ on some input. Consequently, to achieve an error at most $\epsilon$, we have to retrain the model at least every $(4\epsilon)^2$ insertions. For any constant error $\epsilon$, this implies $\frac{n}{(4\epsilon)^2}=O(n)$ model retraining is needed when inserting $n$ data points. Model retraining for many practical choices costs $O(n)$ (to go over the data at least once), so that amortized insertion cost, i.e., insertion \textit{cost per insertion}, must be at least $O(n)$. This is significantly larger than non-learned methods, e.g., for indexing B-trees support insertions in $O(\log n)$. 


Nonetheless, the $\sqrt{k}$ barrier (and consequently a heavy insertion cost) can be avoided, as is often done in practice,  by \textit{partial} retraining. A common example is arranging a set of models in a tree structure and retraining parts of the tree structure as new data is inserted. This avoids a full retraining every $O(\epsilon^2)$ insertions, but makes smaller necessary adjustments throughout that are cheap to make. Thus, Theorem~\ref{lemma:sqrt_n_lower_bound} provides a theoretical justification for many practical design choices in existing work~\cite{ding2020alex, zeighami2023neurosketch, galakatos2019fiting} that partition the space and train multiple models, utilizing data structures built around multiple models to perform operations. \revision{We also note that such approaches often come with repartitioning and tree balancing as new data is inserted to adjust the created partitions after observing new points. Indeed such repartitioning is also necessary in the presence of insertions. Given that for a fixed set of partitions the number of points per partition will grow linearly in data size, Theorem~\ref{lemma:sqrt_n_lower_bound} can be used to show the error 
per partition will remain large unless partitions are recreated and adjusted as new points are observed}.

\revision{The error in Theorem~\ref{lemma:sqrt_n_lower_bound} is independent of total data size, $j$, and only depends on $k$. This is because we make no assumptions on model capacity, and consequently, when the model is trained on the dataset, $\mD^i$ of size $i$, the training error can be zero. Thus, error on $\mD^j$ only depends on how well the trained model on $\mD^i$ generalizes to $\mD^j$, which, intuitively, only depends on the difference between $\mD^i$ and $\mD^j$. Theorem~\ref{lemma:sqrt_n_lower_bound} quantifies this difference in terms of $k$.}


\textbf{Analysis Framework Overview}. In light of Theorem~\ref{lemma:sqrt_n_lower_bound} and existing practical modeling choices that use a set of models to perform an operation, analyzing database operations in the presence of insertion can be divided into two components: (1) how well a model can learn a set of observations (\textit{the modeling problem}), and (2) how a set of models can be used to perform operations in the presence of insertions (\textit{the model utilization problem}). Our framework allows studying the two separately, as discussed next. 

\vspace{-0.2cm}
\subsection{The Modeling Problem}
\vspace{-0.1cm}
Our analysis is divided into studying the problem of modeling and the problem of model utilization. 
We introduce the notion of \textit{distribution learnability} to abstract away the modeling problem when studying the utilization problem. Roughly speaking, if a distribution class is \textit{distribution learnable}, we can use observations from the class to model their distribution operation functions \textit{well}. In other words, if a distribution class is distribution learnable, we have a solution to the modeling problem, and thus, we can focus on the model utilization problem. Meanwhile, the modeling problem is reduced to showing distribution learnablity. In this section, we define \textit{distribution learnability} and discuss how we can prove a distribution class is distribution learnable.

\subsubsection{Defining Distribution Learnability}\label{sec:def_dist_learn}

A distribution class is distribution learnable if there exists an algorithm that returns a good model of the data distribution given an observed dataset. Formally,
\begin{definition}\label{def:dist_learnable}
    A distribution class $\mathfrak{X}$, is said to be \textit{distribution learnable for an operation $f$}, $f\in \{r, c\}$,  with parameters $\mathcal{T}_n^{\mathfrak{X}}$, $\mathcal{S}_n^{\mathfrak{X}}$ and $\mathcal{B}_n^{\mathfrak{X}}$, if for any $\chi\subseteq\mathfrak{X}$, there exists an algorithm that takes a set of observations, $\mD\sim\chi$, of size $n$ as input and returns a model $\hat{f}$ such that:
    \vspace{-0.3cm}\begin{itemize}
        \item (\textit{Accuracy}) If $\mD$ is sampled from $\chi$, for some  $\chi\subseteq\mathfrak{X}$, we have that $$\mathds{P}_{D\sim\chi}[\normx{f_\chi-\hat{f}}_\infty\geq \epsilon]\leq \varkappa_1e^{-\varkappa_2(\frac{\epsilon}{\sqrt{n}}-1)^2},$$ For any $\epsilon \geq \sqrt{n}$ and universal constants $\varkappa_1,\varkappa_2>0$; 
        \vspace{-0.2cm}\item (\textit{Inference Complexity}) It takes $\mathcal{T}_n^{\mathfrak{X}}$ number of operations to evaluate $\hat{f}$ and  space $\mathcal{S}_n^{\mathfrak{X}}$ to store it; and
        \vspace{-0.2cm}\item (\textit{Training Complexity}) Each call to the algorithm costs $\mathcal{B}_n^{\mathfrak{X}}$ amortized number of operations.
    \end{itemize}
\end{definition}\vspace{-0.2cm}
That a distribution class is distribution learnable for operation $f$ means that observations from the distribution class can be used to model the expected value of $f$ to a desired accuracy, and that distribution dependent parameters $\mathcal{T}_n^{\mathfrak{X}}$, $\mathcal{S}_n^{\mathfrak{X}}$ and $\mathcal{B}_n^{\mathfrak{X}}$, characterize the computational complexity of the modeling (amortized number of operations is total number of operations divided by $n$, so $n\mathcal{B}_n^{\mathfrak{X}}$ is total number of operations). 
We make two remarks regarding the definition. 

\begin{remark}
The accuracy requirement for distribution learnability is defined so that, with high probability, the model error is at most $O(\sqrt{n})$. This is due to Theorem~\ref{lemma:sqrt_n_lower_bound}, which shows the expected generalization error, after $n$ insertions, will be $\Omega(\sqrt{n})$ as dataset changes because of insertions and irrespective of modeling accuracy. Thus, having modeling error lower than $O(\sqrt{n})$ will not improve the generalization error, but will increase inference complexity (larger models will be needed to improve accuracy). Meanwhile, due to the inherent $\Omega(\sqrt{n})$ error, an extra modeling error of $\sqrt{n}$ only increases generalization error by a constant factor, thus not changing any of our results asymptotically.

\revision{We also note that, given that Def.~\ref{def:dist_learnable} allows modeling errors to occur, using distribution learnability to return exact results (i.e., for indexing and sorting) requires designing fallback strategies to ensure correctness even in the presence of modeling error. This is done in our results in Sec.~\ref{sec:using_oracles}, where model outputs are adjusted by some lightweight non-learned method (e.g., exponential search for indexing and merge sort for sorting) to ensure correctness. On the other hand, for cardinality estimation, where the goal is to obtain fast estimates and errors can be tolerated in practice, our results in Sec.~\ref{sec:using_oracles} use the modeling error formulation in Def.~\ref{def:dist_learnable} to present methods that answers queries with guaranteed bounds on error.}
\end{remark}

\begin{remark}
Distribution learnability for the distribution class, $\mathfrak{X}$ is defined so that we can characterize the computational complexity of modeling data from $\mathfrak{X}$. Such characterization is important because different modeling choices are beneficial for different distributions. For instance, a linear model may be sufficient to model data from uniform distribution but not for a Gaussian distribution. The definition allows us to distinguish between simple distribution classes where we can create models that are fast to evaluate (e.g., linear models for uniform distribution), from more complex distribution classes that may need more complex models with higher runtime and space complexity (e.g., neural networks for complex distributions). This is done through parameters $\mathcal{T}_n^\mathfrak{X}$, $\mathcal{S}_n^\mathfrak{X}$ and $\mathcal{B}_n^\mathfrak{X}$. 
\end{remark}

\vspace{-0.1cm}
\subsubsection{Proving Distribution Learnability}\label{sec:prove_dist_learn}
\vspace{-0.1cm}
For a distribution class, $\mathfrak{X}$, to be distribution learnable for an operation $f$, we need to be able to model distribution operation functions in that class using some model class $\mathcal{F}$. Intuitively, $\mathcal{F}$ needs to have \textit{enough {representation power}} to model distributions in $\mathfrak{X}$, and we need to be able to effectively optimize over $\mathcal{F}$ to find the good representations given an input (i.e., $\mathcal{F}$ is \textit{optimizable}). The following theorem shows that if these two properties are true, then the distribution class is indeed distribution learnable. 
For the sake of space, we only state our results here informally (formal statement is in Sec.~\ref{appx:sec:formal_func_appx}), as a formal statement requires making \textit{enough representation power} and \textit{opitimizability} concrete, which diverts from our main discussion. 
\begin{theorem}[Informal]\label{thm:generic_oracle_construction}
    \revision{Let $\mathfrak{X}$ be a distribution class whose operation functions belong to some function class $\mathcal{G}$. That is, $f_\chi\in \mathcal{G}$ for all $\chi\in\mathfrak{X}$, for an operation $f$. Assume another function class, $\mathcal{F}$, has enough representation power to represent $\mathcal{G}$, and is optimizable}. Then, $\mathfrak{X}$ is distribution learnable for operation $f$. 
\end{theorem}
\vspace{-0.1cm}
Theorem~\ref{thm:generic_oracle_construction} can be broadly used to translate  function approximation results to distribution learnability. For instance, Taylor's theorem shows that infinitely differentiable functions can be approximated by polynomials to arbitrary accuracy (i.e., polynomials have enough representation power to represent infinitely differentiable functions), and the exchange algorithm~\cite{powell1981approximation} shows that we can find the best polynomial approximating a function (i.e., shows optimizability for polynomials). These together with Theorem~\ref{thm:generic_oracle_construction} imply that distributions with infinitely differentiable operation functions are distribution learnable. 
Nonetheless, the time complexity of function approximation is important when deciding what function class to choose for modeling purposes in database applications. For instance, the exchange algorithm, although converges, can take too long to find polynomials that model functions with a desired accuracy~\cite{powell1981approximation}. Our next result uses Theorem~\ref{thm:generic_oracle_construction} to show distribution learnability using piecewise linear and piecewise constant models that show better time/space complexity. We first discuss learnability for rank operations.

\begin{table}[t]
    \centering
    \vspace{-0.2cm}
    \begin{tabular}{c l l l}
        \toprule
        \textbf{Distribution class} & $\mathbf{\mathcal{T}_{n}^{\mathfrak{X}}}$ & $\mathbf{\mathcal{S}_{n}^{\mathfrak{X}}}$ &  $\mathbf{n\mathcal{B}_{n}^{\mathfrak{X}}}$  \\\midrule
        $\mathfrak{X}_\rho$ & 1 & $\rho\sqrt{n}\log n$ & $\rho\sqrt{n}\log n$  \\
        $\mathfrak{X}_l$ & $\log l$ & $l\log n$ & $n\log n$  \\
        $\mathfrak{X}_c$ & 1 & 1 & $1$  \\
        \bottomrule
    \end{tabular}
    \vspace{-0.2cm}
    \caption{Asymptotic complexities of some distribution learnable classes for rank function defined in Lemma~\ref{lemma:specific_oracle_construction}}
    \label{tab:specific_dist_classes}
\end{table}
\vspace{-0.2cm}
\begin{lemma}\label{lemma:specific_oracle_construction}
    Let $\mathfrak{X}_\rho$ be the set of distributions with p.d.f bounded by $\rho$, $\mathfrak{X}_l$ the set of distributions with piecewise linear c.d.f with at most $l$ pieces and $\mathfrak{X}_{c}$ a distribution the c.d.f of which can be stored and evaluated in constant time. $\mathfrak{X}_\rho$, $\mathfrak{X}_l$, $\mathfrak{X}_c$ are distribution learnable for rank operation with parameters shown in Table~\ref{tab:specific_dist_classes}.
\end{lemma}
\vspace{-0.2cm}

Lemma~\ref{lemma:specific_oracle_construction} presents results for multiple distribution classes. $\mathfrak{X}_\rho$ formulates a realistic scenario (experimentally shown by \citep{zeighami2023distribution}). $\mathfrak{X}_c$ shows the ideal scenario for learned models, where the data distribution is easy to model, and is included to show a best-case scenario for our results when using learned models. Piece-wise linear models have been used for the purpose of indexing~\citep{ferragina2020pgm,galakatos2019fiting}, and $\mathfrak{X}_l$ is included to study their theoretical properties for the distribution class where they are well suited. Next, consider distribution learnability for cardinality operation.\vspace{-0.1cm}
\begin{lemma}\label{lemma:specific_oracle_construction_ce}
    Let $\mathfrak{X}_\rho$ be the set of distributions for which the distribution cardinatliy function has gradient bounded by $\rho$, and let $\mathfrak{X}_{c}$ be a countable set of $c$ distributions for which distribution cardinatliy function can be stored and evaluated in constant time. $\mathfrak{X}_\rho$ and $\mathfrak{X}_c$ are distribution learnable for cardinality estimation where the same parameters as Table~\ref{tab:specific_dist_classes} hold for $\mathfrak{X}_c$. For $\mathfrak{X}_\rho$, we have $\mathcal{B}_{n}^{\mathfrak{X}}$ and $\mathcal{S}_{n}^{\mathfrak{X}}$ as $O(\sqrt{2d}(\rho\sqrt{n})^{2d}\log n)$, while $\mathcal{T}_{n}^{\mathfrak{X}}=O(1)$.
\end{lemma}\vspace{-0.1cm}
As before, we have included $\mathfrak{X}_{c}$ to show a best-case scenario for learned models. Nonetheless, cardinality estimation is a problem in high dimensions where modeling is difficult. The exponential behavior in Lemma~\ref{lemma:specific_oracle_construction_ce} for $\mathfrak{X}_\rho$ is required for different modeling choices, including neural networks \citep{petersen2018optimal, yarotsky2018optimal}. To reduce complexity, stricter assumptions on data distribution are often justified. For example, attributes may be correlated and only fall in a small part of the space. A common assumption using neural networks is that data is supported on a low dimensional manifold \cite{pope2021intrinsic}, which together with results showing that neural networks can approximate data on low dimensional manifolds well \citep{chen2019efficient}, yields that neural networks can avoid space complexity exponential in dimensionality. This is an active area of research orthogonal to our work, and our results show how learned database operations can benefit from such approximation theoretic results as they become available. 

\vspace{-0.2cm}
\subsection{The Model Utilization Problem}\label{sec:utilization_def}
\vspace{-0.2cm}
Our results in Sec.~\ref{sec:using_oracles} thoroughly discuss how learned models can perform different database operations for distribution learnable classes. Here, we provide a brief overview of the general methodology and state required definitions. 

Typical methods used in practice to perform database operations partition the domain and model different parts of the domain separately. 
Each partition can be denoted by a set $R$ of the space it covers. The model in each partition can be seen as a model of the conditional distribution of the data, where the original data distribution is conditioned on the set $R$. As such, to effectively model the data in a partition, we need to be able to model the conditional distribution for the partition. This means not only the original data distribution, but also the conditional data distributions need to be distribution learnable. We formalize our notion of conditional distribution to be able to formalize this statement.  

Let $\mathcal{R}$ be a set s.t. $\mathcal{R}\subset2^{\mathcal{D}}$ (i.e., $\mathcal{R}$ is a set of subsets of the data domain). Then, for any $R\in \mathcal{R}$ with $\mathds{P}_{X\sim\chi}(X\in R)>0$, we define $\chi|R$ as the data distribution with c.d.f $F_{\chi|R}(\vx)=\mathds{P}_{X\sim\chi}(X\leq \vx|X\in R)$.  In this paper, unless otherwise stated, $\mathcal{R}$ is the set of axis-parallel rectangles, where $R=(\vr_{min}, \vr_{max})$ with $\vr_{min}, \vr_{max}\in [0, 1]^d$ define two corners of the hyper rectangle. We define the \textit{normalized} conditional distribution, $\bar{\chi|R}$, as the distribution with c.d.f ${F}_{\bar{\chi|R}}(\vx)=F_{\chi|R}((\vr_{max}-\vr_{min})\vx+\vr_{min})$. The normalization scales the domain of the conditional distribution back to [0, 1]$^d$, and helps standardize our modeling discussion. We define the closure of a distribution class $\mathfrak{X}$, denoted by $\bar{\mathfrak{X}}$,  as the set $\{\bar{\chi|R}, \forall \chi\in \mathfrak{X}, R\in\mathcal{R}\}$. That is, $\bar{\mathfrak{X}}$ contains not only $\mathfrak{X}$ but all the other distributions obtained by distributions $\chi\in\mathfrak{X}$ conditioned under sets $R\in\mathcal{R}$. Often, we need the distribution class $\bar{\mathfrak{X}}$, and not only $\mathfrak{X}$, to be distribution learnable. $\bar{\mathfrak{X}}$ and $\mathfrak{X}$ can be (but not necessarily are) the same set. An example is the uniform distribution, where conditioning the distribution over any interval yields another uniform distribution over the interval.

%% file: results.tex
\vspace{-0.1cm}
\subsection{Indexing Dynamic Data}
\vspace{-0.15cm}
We show the following result for dynamic indexing.\vspace{-0.1cm}
\begin{theorem}\label{thm:dynamic_indexing_oracle}
    Suppose $\mD\sim\chi$ for $\chi\subseteq\mathfrak{X}$ for some distribution class $\mathfrak{X}$ with $TV(\chi)\leq \delta$, and that $\bar{\mathfrak{X}}$ is distribution learnable. 
    There exists a learned index into which the $n$ data points of $\mD$ can be inserted in $O(\mathcal{T}_n^{\mathfrak{X}}\log\log n+\log \delta\sqrt{n}+\mathcal{B}_n^{\mathfrak{X}}\log^2\log n)$ expected amortized time, that can be queried in $O(\mathcal{T}_n^{\mathfrak{X}}\log\log n+\log \delta\sqrt{n})$ expected time and takes space $O(n\log n+\sum_{i=0}^{\log\log n}n^{1\shortminus2^{\shortminus i}}\mathcal{S}_{n^{2^{\shortminus i}}}^\mathfrak{X})$. 
\end{theorem}\vspace{-0.15cm}
The term $\mathcal{T}_n^{\mathfrak{X}}\log\log n$ is due to making $\log\log n$ calls to the distribution model, and $\mathcal{B}_n^\mathfrak{X}$ roughly refers to the need to rebuild a model every $n$ insertions. For example, without distribution shift (i.e., $\delta=0$), one can answer queries and perform insertions with $O(\log\log n)$ model calls, while every $n$ insertions incurs extra $\mathcal{B}_n^\mathfrak{X}$ cost for model rebuilding. 

Distribution shift increases both insertion and query time by $O(\log \delta\sqrt{n})$. In the worst case, having $\delta=1$, we recover the traditional $O(\log n)$ insertion and query time. That is, our results show no gain from modeling when distribution shift is too severe. This is as expected. If data distribution changes too much, one cannot use the current knowledge of data distribution to locate future elements. By systematically handling the distribution shift, we show that a learned method can provide robustness in such scenarios.

The data structure that achieves the bound is a tree structure with a distribution model used in each node to find the node's child to traverse given a query or insertion. The structure can be thought of as a special case of Alex~\cite{ding2020alex}, with specific tree height, fanount and split mechanism to ensure the desired gaurantees. All elements are stored at leaf nodes, and the traversal to the leaf nodes is similar to B-trees but using learned models to choose the child. 
Using Lemma~\ref{lemma:specific_oracle_construction} we can specialize Theorem~\ref{thm:dynamic_indexing_oracle} for specific distribution classes.
\begin{corollary}\label{corollary:indexing_specific_construction}
    Let $\mathfrak{X}_{\rho_1,\rho_2}$ be the class of distributions with bounded p.d.f. \revision{That is, for all $\chi\in\mathfrak{X}_{\rho_1,\rho_2}$ and denoting by $g_\chi$ the p.d.f of $\chi$, we have $0<\rho_1\leq g_\chi(x)\leq \rho_2<\infty, \;\forall x$}. Suppose $\mD\sim\chi$ for $\chi\subseteq\mathfrak{X}$ for some distribution class $\mathfrak{X}\subseteq \mathfrak{X}_{\rho_1,\rho_2}$ with $TV(\chi)\leq \delta$. There exists a learned index that supports insertions in $O(\log\log n+\log \delta\sqrt{n})$ expected amortized time, queries in $O(\log\log n+\log \delta\sqrt{n})$ expected time and takes space $O(\frac{\rho_1}{\rho_2}n\log n)$. 
\end{corollary}\vspace{-0.2cm}
Corollary~\ref{corollary:indexing_specific_construction} shows a learned index that performs insertions and answers queries in $O(\log\log n+\log(\delta n))$, while non-learned methods take $O(\log n)$. Thus, when distribution shift is not severe, a learned method can outperform non-learned methods, while large distribution shift ($\delta=1$) leads to same bounds as non-learned methods. Corollary~\ref{thm:dynamic_indexing_oracle} strictly generalizes results in \cite{zeighami2023distribution} to the setting with insertions and data distribution change.

\revision{We note that one can consider the data structure used in the proof of Theorem~\ref{thm:dynamic_indexing_oracle} (and consequently Corollary~\ref{corollary:indexing_specific_construction}) as a method for switching between learned and non-learned indexes when the distribution shift grows. Recall that proof of Theorem~\ref{thm:dynamic_indexing_oracle} uses a tree of learned models as an index. When there is no distribution shift, this tree is traversed only using learned models. However, when the distribution shift is large, the tree structure needs to be adjusted frequently (i.e., nodes are split frequently), and these adjustments are stored in a non-learned data structure. Thus, when the distribution shift is large, the tree traversal becomes more reliant on non-learned data structures.}

\vspace{-0.3cm}
\subsection{Cardinality Estimation}\vspace{-0.2cm}
For cardinality estimation, designing learned models that answer queries with arbitrary accuracy is more challenging due to the high dimensionality of the problem. The curse of dimensionality is a well-understood phenomenon for non-learned methods, leading to approaches that take space exponential in dimensionality~\citep{graham2012synopses, wei2018tight}. We first show that this is not the case when using learned models if an error of $\Omega(\sqrt{n})$ is tolerable. 
\vspace{-0.1cm}
\begin{theorem}\label{thm:dynamic_ce_hd_oracle}
    Suppose $\mD\sim\chi$ for $\chi\subseteq\mathfrak{X}$ for a distribution learnable class $\mathfrak{X}$ with $TV(\chi)\leq \delta$. 
    There exists a learned cardinality estimator that answers queries with expected error $\epsilon$ for $\epsilon=\Omega(\sqrt{n})$ supports insertions in $O(\max\{\frac{\delta \sqrt{n}}{\epsilon}, 1\}\mathcal{B}_n^{\mathfrak{X}})$, queries in $O(\mathcal{T}_n^{\mathfrak{X}})$ and takes space $O(\mathcal{S}_n^{\mathfrak{X}})$. 
\end{theorem}
\vspace{-0.1cm}

Theorem~\ref{thm:dynamic_ce_hd_oracle} states that we can use a distribution model to answer queries for any expected error $\Omega(\sqrt{n})$. Consequently, when we can effectively model a data distribution, we can answer queries to accuracy at least $\sqrt{n}$ without having an exponential space blowup. Comparing this with random sampling, and more broadly $\epsilon$-approximations, that need at least $\sqrt{n}\log^{d-1}(\sqrt{n})$ data samples to answer queries with accuracy $\sqrt{n}$ \citep{wei2018tight, matouvsek2015combinatorial}, we show a clear advantage to learned models over such non-learned methods in this accuracy regime. 

\vspace{-0.1cm}
Theorem~\ref{thm:dynamic_ce_hd_oracle} uses a single distribution model that is periodically retrained with insertion. The frequency of retraining depends on distribution shift. If $\delta\leq \frac{1}{\sqrt{n}}$, the error caused by distribution shift is on a similar scale as error due to randomness. Thus, the distribution shift does not significantly affect insertion time. On the other hand, in the worst case when $\delta=1$, we need to retrain the model every $\frac{1}{\epsilon}$ insertions, which can be significant depending on retraining cost.


Error of $\Omega(\sqrt{n})$ is not necessarily too large. Indeed, expected query answer for a fixed query with probability $p$ is $n\times p$, so error, relative to the expected query answer is $O(\frac{\sqrt{n}}{np})=O(\frac{1}{\sqrt{n}})$ and goes to zero as data size increases. Nonetheless, one may wish to answer queries more accurately. Below, we discuss how to achieve this in one dimension. Appendix~\ref{appx:hd_ce_oracle_anyaccuracy}, presents Lemma~\ref{lemma:hd_ce_oracle_anyaccuracy} that shows how ideas in one dimension can be extended to high dimensions, but nevertheless, only achieves space complexity exponential in dimensionality, similar to non-learned methods.

\textbf{Arbitrary Accuracy in One Dimension}. In one dimension, we show the following is possible using learned models. 
\vspace{-0.1cm}
\begin{theorem}\label{thm:dynamic_ce_1d_oracle}
    Suppose $\mD\sim\chi$ for $\chi\subseteq\mathfrak{X}$ for a distribution class $\mathfrak{X}$ with $TV(\bar{\chi})\leq \delta$, and that $\mathfrak{X}$ is distribution learnable. There exists a learned cardinality estimator that answers queries with expected error $\epsilon$ for any $\epsilon>0$ supports insertions in $O(\max\{delta\epsilon, \epsilon^2\}\mathcal{B}_{\epsilon^2}^\mathfrak{X})+\log n)$, queries in $O(\log n+\mathcal{T}_{\epsilon^2}^\mathfrak{X})$ and takes space $O(\frac{n}{\epsilon^2}\mathcal{S}_{\epsilon^2}^\mathfrak{X}+\frac{n}{\epsilon^2}\log n)$.
\end{theorem}
\vspace{-0.2cm}
Theorem~\ref{thm:dynamic_ce_1d_oracle} shows that we can effectively answer queries to any accuracy in one dimension using learned models. Importantly, the result shows that if the data distribution can be modeled space-efficiently (e.g., whenever $\mathcal{S}_{\epsilon^2}^\mathfrak{X}\leq \log n$), then a learned approach outperforms non-learned (and worst-case optimal) method discussed in \citep{wei2018tight} that takes space $O(\frac{n}{\epsilon}\log n)$ to answer queries with accuracy $\epsilon$. 

\vspace{-0.1cm}
The learned model that achieves the bound in Theorem~\ref{thm:dynamic_ce_1d_oracle} uses a combination of materialized answers and model estimates to answer queries. Given that a model can be at best accurate to $\sqrt{n}$ if the dataset contains $n$ points, the algorithm divides up the data domain into $\frac{n}{\epsilon^2}$ intervals, each containing $\epsilon^2$ points, so that a model for each interval will have accuracy $\epsilon$. Meanwhile, the algorithm materializes query answers that span multiple intervals so that errors do not accumulate when answering such queries.  The materialization is done through a B-tree like structure, where each node stores the exact number of points inserted into it. Because, in our construction, we build several models each for a subset of the space, it is not enough that the total variation between the distributions is bounded, but also that the total variation after conditioning is bounded ($\bar{\chi}$ is the closure of $\chi$ under conditioning defined in Sec.~\ref{sec:utilization_def}).

\vspace{-0.2cm}
\subsection{Sorting}
\vspace{-0.2cm}
Sorting involves only a fixed array, while the operations we discussed so far consider a dataset that changes due to insertions. Our discussion here shows that the distribution learnability framework can be beneficial for analyzing learned database operations beyond insertions.

To see why our framework applies to learned sorting, first recall the existing learned sorting algorithm of \citet{kristo2020case}, which sorts an array by first sampling a subset of the array, learning a model to predict the correct location using the sample (the sample is sorted by an existing algorithm for the purpose of training), and then using the learned model to predict the item locations in the original array. Here, similar to the case of learned operations with insertions, the problem isn't (only) how well we can learn a model, but also how well a model learned from a sample of the array will generalize to the complete array. Thus, we need to both study a modeling problem and a model utilization problem, and we can do so using distribution learnability. Finally, since a function that sorts an array is the rank function, our discussion on distribution learnability for the rank operation already covers the modeling problem. 

Before stating our results, we also note that one can sort a fixed array by iteratively inserting its element into a learned index. Thus, Theorem~\ref{thm:dynamic_indexing_oracle} has already provided a method for sorting an array using machine learning. Our result below presents another learned method for sorting an array. This is analogous to how both B-trees and merge sort can be used to sort a fixed array. The result below can be seen as a means of extending merge sort with machine learning. \vspace{-0.1cm}
\begin{theorem}\label{thm:sorting_oracle}
    Suppose an array consists of $n$ points sampled i.i.d. from a distribution learnable class $\mathfrak{X}$.  There exists a learned sorting method that sorts the array in $O(\mathcal{T}_{\sqrt{n}}^\mathfrak{X}n\log\log n+\sqrt{n}\mathcal{B}_{\sqrt{n}}^\mathfrak{X}+\sum_{i=0}^{\log\log n}n^{1\shortminus\frac{1}{2}(\frac{4}{5})^{i}}\mathcal{B}_{n^{\frac{1}{2}(\frac{4}{5})^{i}}}^\mathfrak{X})$ and space $O(\mathcal{S}_{\sqrt{n}}^\mathfrak{X}+n\log n)$
\end{theorem}
\vspace{-0.2cm}
We note that $\sqrt{n}\mathcal{B}_{\sqrt{n}}^\mathfrak{X}+\sum_{i=0}^{\log\log n}n^{1\shortminus\frac{1}{2}(\frac{4}{5})^{i}}\mathcal{B}_{n^{\frac{1}{2}(\frac{4}{5})^{i}}}^\mathfrak{X})$ is $O(n)$ if model training takes linear in data size, so that the runtime is $O(\mathcal{T}_{\sqrt{n}}^\mathfrak{X}n\log\log n)$. This is independent of training time, $\mathcal{B}_n^\mathfrak{X}$, because, due to sampling, training is done on much smaller arrays than the original data. Thus, Theorem~\ref{thm:sorting_oracle} provides a time complexity for sorting similar to Theorem~\ref{thm:dynamic_indexing_oracle}, both showing that for efficient modeling choices, one can sort an array in $O(n\log\log n)$ model calls.  

The algorithm that achieves the bound in Theorem~\ref{thm:sorting_oracle} is similar to \citet{kristo2020case}, which first samples a subset of the original array, uses it to build a distribution model, and uses the model to sort the original array. Due to modeling errors, the resulting attempt using the model will only be a partially sorted array. Unlike \citet{kristo2020case} that uses insertion sort to fully sort the partially sorted array, we use a merge sort like approach to recursively sort the array. This is because, to reduce the asymptotic complexity below $O(n\log n)$, the sample needs to be of size $o(n)$. However, the generalization error of a model trained on a sample of size $o(n)$ would be too large to allow insertion sort to be effective. Partitioning the partially sorted array and recursively sorting each portion allows us to sort the array while performing $O(\mathcal{T}_{\sqrt{n}}^{\mathfrak{X}}n\log \log n)$ operations.


Finally, the lower bound of Theorem~\ref{lemma:sqrt_n_lower_bound} does not apply to sorting, and a natural question is if it is possible to do better than $O(n\log\log n)$. The following result shows that under stronger assumptions on data distribution, this is possible. 

\vspace{-0.1cm}
\begin{theorem}\label{thm:sorting_oracle_accurate}
        Suppose an array consists of $n$ points sampled i.i.d. from a distribution $\chi$, and assume we have a model $\hat{r}$ s.t., $\normx{\hat{r}-r_\chi}_\infty\leq \epsilon_0$, that can be evaluated in $\mathcal{T}^\chi$ and takes space $\mathcal{S}^\chi$. There exists an algorithm that sorts the array in $O(\mathcal{T}^\chi n\log\epsilon_0)$ taking space $O(\mathcal{S}^\chi+n)$.
\end{theorem}
\vspace{-0.2cm}

The theorem shows if we know the data distribution very accurately, then we can sort the data very efficiently.  $\mathcal{T}^\chi$ can be $O(1)$, e.g., if the data c.d.f was a polynomial, so that we can sort an array in $O(n)$. This is because the data distribution provides a good indicator of the location of the item in the sorted array. The algorithm that achieves this can be seen as a special case of merge sort, where instead of dividing the array into 2, we divide the array, using $\hat{r}$, to up to $n$ groups, and recursively sort each.

The difference between Theorems~\ref{thm:sorting_oracle_accurate} and~\ref{thm:sorting_oracle} is how accurate of a model of data distribution we have access to. Theorem~\ref{thm:sorting_oracle} effectively assumes that data distribution can only be modeled to accuracy $O(\sqrt{n})$, which is too large to allow fixing model errors in sorting with a single pass over the array. On the other hand, Theorem~\ref{thm:sorting_oracle_accurate} assumes the model is correct to within a constant accuracy. As a result, a single iteration over the partially sorted array fixes any potential inversions and yields the $O(n)$ complexity. Nonetheless, knowing the data distribution to a constant accuracy can be impractical, because it requires further knowledge about the data distribution beyond merely observing its samples.

%% file: conclusion.tex
\vspace{-0.4cm}
\section{Conclusion and Future Work}
\vspace{-0.2cm}
We have presented a thorough theoretical analysis of learned indexing and cardinality estimation in the presence of insertions from a possibly changing data distribution. Our results characterize learned models' performance, and show when and why they can outperform their non-learned counterparts. We have developed the distribution learnability analysis framework that provides a systematic tool for analyzing learned database operations. Our results enhance our understanding of learned database operations and provide the much-needed theoretical guarantees on their performance for robust practical deployment. 

\revision{We believe our theoretical tools will pave the way for a broader theoretical understanding of various learned methods. Nonetheless, there are several aspects that require further research to allow for a broader applicability of our results to real-world databases. First, we have established distribution learnability for distributions with bounded p.d.f and piecewise-linear c.d.f, but demonstrating distribution learnability for a broader range of real-world data distributions is needed to cover a more comprehensive set of real-world data distributions. There are two aspects that require further research. There is a need to identify and formally characterize the distribution classes from which real-world datasets originate. This task is challenging, especially for high-dimensional data. For instance, in the case of images, it is commonly believed that they lie on a low-dimensional manifold. It is essential to validate if such assumptions also hold true for real-world tabular datasets.  Moreover, we need to establish distribution learnability for such distribution classes using appropriate modeling choices. The best modeling choice may vary depending on the distribution class, necessitating further research to determine the most effective modeling choices for specific distribution classes. Other future work includes incorporating deletions and updates, where we believe statistical tools developed here can be utilized, but formalizing the notion of data distribution in the presence of deletions/updates, and the relationship between insertions and deletions/updates require further research}. Finally, considering query distribution, and analyzing other database operations are other future directions.  

%% file: rel_work.tex
A large and growing body of work has focused on using machine learning to speed up database operations, among them, learned indexing \citep{galakatos2019fiting, kraska2018case, ferragina2020pgm, ding2020alex}, learned cardinality estimation \citep{kipf2018learned, wu2021unified,hu2022selectivity,yang2019deep,yang2020neurocard, lu2021pre, negi2021flow} and learned sorting \citep{kristo2020case}. Most existing work focus on improving modeling choices, with various modeling choices such as neural networks \citep{zeighami2023neurosketch, kipf2018learned, kraska2018case}, piece-wise linear approximation \citep{ferragina2020pgm}, sum-product networks \citep{hilprecht2019deepdb} and density estimators \citep{ma2019dbest}. Existing results show significant empirical benefits in static datasets, while performance often deteriorates in dynamic datasets and in the presence of distribution shift \citep{wang2021we, wongkham2022updatable}. Our theoretical results help explain such observations and provide a theoretical framework for analysis of the operations under different modeling choices. 

On the theory side, no existing study meaningfully characterizes performance of learned models in the dynamic setting or studies learned sorting. In the static setting, \citep{zeighami2023distribution, ferragina2020learned} study query time of learned indexing. \citet{ferragina2020learned} shows learned models can provide constant factor improvements under an assumption on the distribution of the gap between observations, and \citet{zeighami2023distribution} shows a learned model can answer queries in $O(\log\log n)$ query time if the p.d.f of data distribution is non-zero and bounded. Our results strictly generalize the latter to the dynamic setting, in the presence of insertions from a possibly changing distribution, and also show that more generally, $O(\mathcal{T}_n^{\mathfrak{X}}\log\log n+\log\delta n)$ query time is possible for any distribution learnable class $\mathfrak{X}$. Moreover, \citet{zeighami2023neurosketch} presents a special case of our Theoerem~\ref{thm:dynamic_ce_hd_oracle} for cardinality estimation on static datasets for distributions where the operation distribution function is Lipschitz continuous. Our result strictly generalizes \citet{zeighami2023neurosketch} to the dynamic setting with distribution change and any distribution learnable class $\mathfrak{X}$. 
Orthogonal to our work, \citet{zeighami2023towards} study lower bounds on the model size needed to perform various database operations with a desired accuracy and \cite{hu2022selectivity, agarwala2021one} study the number of training samples needed to achieve a desired accuracy for different database operations. 

Finally, we draw a broader connection between our work, learned indexing and interpolation search. A large body of early work focused on interpolation search \citet{peterson1957addressing, perl1978interpolation, perl1977understanding, yao1976complexity, mehlhorn1993dynamic}, proposed by \cite{peterson1957addressing} which uses linear interpolation to estimate the location of a query in an array. It has been shown that this search algorithm achieves $O(\log\log n)$ query time on uniformly distributed arrays \cite{yao1976complexity, perl1977understanding}, with extensions to cover \textit{smooth} distribution classes and dynamic data in \cite{andersson1993dynamic, mehlhorn1993dynamic}. Indeed, interpolation search can be seen as an early example of a model-based search, where linear models are used to estimate item locations.  Using the terminology introduced in this paper and given that uniform distribution is distribution learnable using linear models (c.d.f of the uniform distribution is a linear function), the $O(\log\log n)$ query time can be seen as a special case of our results. Overall, interpolation search can be seen as a special case of learned indexing, where learned indexing allows for more complex data-driven modeling choices that can be useful for a broader class of data distributions.

%% file: setup.tex
We are interested in performing database operations on a possibly changing dataset. We assume data records are $d$-dimensional points in the range $[0, 1]$ (otherwise, the data domain can be scaled and shifted to this range). We either consider the setting when $n$ data points are inserted one by one into the dataset, or that we are given a fixed set of $n$ data points. We refer to the former as the dynamic setting and the latter as the static setting. We define $\mD^i$ as the dataset $\mD^i\in [0, 1]^{i\times d}$, i.e., a dataset consisting of $i$ records inserted so far and in $d$ dimensions with each attribute in the range [0, 1], where $i$ and $d$ are integers greater than or equal to 1. $\mD^n$ is the dataset after the last insertion, and is often denoted as $\mD$. $\mD^{i:j}$ denotes the dataset of points inserted after the $i$-th insertion until the $j$-th (i.e., $\mD^{j}\setminus\mD^{i}$). We use $\bm{D_i}$ to refer to the $i$-th record of a dataset (which is a $d$-dimensional vector) and $D_{i, j}$ to refer to the $j$-th element of $\bm{D_i}$. If $d=1$ (i.e., $\mD$ is 1-dimensional), then $D_i$ is the $i$-th element of $\mD$ (and is not a vector). We study the following database operations. 

\textit{Indexing}. The goal is to use an index to store and find items in a $1$-dimensional dataset. The index supports insertions and queries. $n$ items are inserted into the index one by one. After inserting $k$ items, for any $1\leq k\leq n$, we would like to retrieve items from the dataset based on a query $q\in[0, 1]$. The query is either an exact match query or a range query. An exact match query returns the point in the database that exactly matches the query $q$ (or \texttt{NULL} if there is none) while a range query $[q, q']$ returns all the elements in the dataset that fall in the range $[q, q']$, for $q, q'\in[0, 1]$. 

\textit{Cardinality Estimation}. Used often for query optimization, the goal is to find how many records in the dataset match a range query, where the query specifies lower and upper bound conditions on the values of each attribute. Specifically, the query predicate $\vq=(c_1, ..., c_d, r_1, ..., r_d)$, specifics the condition that the $i$-th attribute is in the interval $[c_i, c_i+r_i]$, for $c_i, r_i\in [0, 1]$. Data records can be inserted into the data set one by one. After the insertion of $k$-th item, for any $1\leq k\leq n$, we would like to obtain an estimate of the cardinality of query $\vq$. We expect that the answers are within error $\epsilon$ of the true answers. That is, if $c(\vq)$ is the true cardinality of $q$ and $\hat{c}(\vq)$ is an estimate, we expect $|c(\vq)-\hat{c}(\vq)|\leq \epsilon$. This guarantee has to hold throughout, and as new elements are inserted in the dataset.

\textit{Sorting}. The goal is to sort a fixed array of size $n$. That is, we are given a one-dimensional array, $\mD$, and the goal is to return an array, $\mD'$, which has the same elements as $\mD$ but ordered so that $D'_i\leq D'_{i+1}$. Unlike indexing and cardinality estimation, sorting assumes a fixed given array that needs to be sorted. Although indexing can often be used to sort an array (e.g., inserting elements one by one into a binary tree sorts a fixed array), we study the problem of sorting more broadly and explore other learned solutions to the problem beyond indexing (e.g., analogous to how merge sort can also be used to sort an array). 

%% file: learnability_through_app_formal.tex
\section{Distribution Learnability Through Function Approximation}\label{appx:sec:formal_func_appx}
We first formalize representation power and optimizablity, and then present a formal statement for Theorem~\ref{thm:generic_oracle_construction}.

\textit{Representation Power}. Consider using a function class $\mathcal{F}$ to approximate another function class $\mathcal{G}$ (e.g., neural networks to approximate real-valued functions). Consider some hyperparameter, $\vartheta$, that controls the representation power and inference complexity in $\mathcal{F}$, and denote by $\mathcal{F}_\vartheta$ is the subset of $\mathcal{F}$ with hyperparameter $\vartheta$. For instance, $\vartheta$ can be the number of learnable parameters of a neural network, the maximum degree of a polynomial, or the number of pieces in a piecewise approximation. In all such cases, larger $\vartheta$ implies better representation power but also higher inference time and/or space complexity. Assume we have access to a \textit{representation complexity} function $\alpha_{\mathcal{F}\rightarrow\mathcal{G}}(\epsilon)$, that given a maximum error $\epsilon$ returns the smallest value of $\vartheta$ such that for any $g\in \mathcal{G}$ there exists an $f\in\mathcal{F}_\vartheta$ with $\normx{f-g}_\infty$. The function $\alpha_{\mathcal{F}\rightarrow\mathcal{G}}(\epsilon)$ determines the required model complexity of $\mathcal{F}$, in terms of $\vartheta$, to represent all  elements of $\mathcal{G}$ with error at most $\epsilon$. For instance, such a function for neural networks approximating real-valued functions will show the minimum number of neural network parameters needed to approximate all real-valued functions to error at most $\epsilon$ with a neural network. We say that a function class, $\mathcal{F}$ has the \textit{representation power} to model $\mathcal{G}$ if there exists a representation complexity function $\alpha_{\mathcal{F}\rightarrow\mathcal{G}}(\epsilon)$ for all $\epsilon>0$. Finally, let $\tau_{\mathcal{F}\rightarrow\mathcal{G}}(\epsilon)$ and $\sigma_{\mathcal{F}\rightarrow\mathcal{G}}(\epsilon)$ respectively be the maximum time and space complexity of performing a model forward pass for functions in $\mathcal{F}_\vartheta$ for $\vartheta=\alpha_{\mathcal{F}\rightarrow\mathcal{G}}(\epsilon)$.

\textit{Optimizability}. 
We say a function class $\mathcal{F}$ is optimizable with an algorithm $\mathcal{A}$ if given any function $h$ and a hyperparameter value $\vartheta$, $\mathcal{A}(h, \vartheta)$  returns an approximately optimal representation of $h$ in $\mathcal{F}_\theta$. Formally, for $\hat{h}=\mathcal{A}(h, \vartheta)$ and if $h^*=\arg\min_{\hat{h}\in \mathcal{F}_\theta}\normx{h-\hat{h}}$, $\normx{h^*-h}_\infty\geq \varkappa \normx{\hat{h}-h}_\infty$ for a constant $\varkappa \leq 1$. Let $\beta(\vartheta)$ be the maximum time complexity of $\mathcal{A}$.  

We note that although optimizability as defined broadly above is sufficient to show distribution learnability, it is not necessary. Here, we discuss two qualifications to the definition that make proving optimizability simpler, specifically for database operations. First, it is only necessary to have optimizability for $h\in f_{\mD}$ for all possible $\mD$ and for a desired operation function $f$ (since we will only use $\mathcal{A}$ to model operation functions). This can simplify the optimizability requirement depending on the operation function considered. For example, when showing opimizability for rank operations, we only need an $\mathcal{A}$ that returns approximately optimal estimates for input functions that are non-decreasing (since all rank functions are non-decreasing).  Second, when $\mathcal{A}$ is used on $h=f_{\mD^n}$, we can allow additive error of $O(\frac{1}{\sqrt{n}})$. That is, we only need to show $\frac{1}{\varkappa}\normx{h^*-h}_\infty +\frac{\varkappa'}{\sqrt{n}}\geq  \normx{\hat{h}-h}_\infty$ for $\varkappa'\geq0$ and $\varkappa\leq 1$ universal constants.


\begin{theorem}\label{thm:generic_oracle_construction_formal}
    Assume a function class, $\mathcal{F}$, is optimizable with an algorithm $\mathcal{A}$, and that $\mathcal{F}$ has enough representation power to represent $\mathcal{G}$. Let $\mathfrak{X}$ be a distribution class with $f_\chi\in \mathcal{G}$ for all $\chi\in\mathfrak{X}$. 
    Then, $\mathfrak{X}$ is distribution learnable with $\mathcal{T}_n^\mathfrak{X}=\tau_{\mathcal{F}\rightarrow\mathcal{G}}(\frac{1}{\sqrt{n}})$, $\mathcal{S}_n^\mathfrak{X}=\sigma_{\mathcal{F}\rightarrow\mathcal{G}}(\frac{1}{\sqrt{n}})$, and $\mathcal{B}_n^\mathfrak{X}=\beta(\alpha_{\mathcal{F}\rightarrow\mathcal{G}}(\frac{1}{\sqrt{n}}))$.
\end{theorem}


%% file: proofs.tex
\section{Proofs}
The high-level idea behind most of our theoretical results is to use the relationship between query answers and distribution properties. Overall, many statistical tools have been developed that relate the properties of an observed dataset to the data distribution (e.g., studying the relationship between sample mean and distribution mean). In statistics, such tools have been used to describe the population using observed samples. Our proofs often use such tools to do the opposite, that is, use the properties of the data distribution to describe observed samples. Indeed, that is the intuition behind learned database operations, that if the data distribution can be efficiently modeled, then it can be used to answer queries about the observed samples (i.e., the database) efficiently. Our proposed distribution learnability framework allows us to state this more formally. It allows us to assume that we can indeed model the data distribution efficiently. Then, the analysis can focus on utilizing statistical tools to characterize the relationship between the observed sample and the data distribution. Having access to an accurate model of the data distribution, we use existing statistical tools to analyze its error. However, a main challenge in the case of learned database operations is to balance accuracy and efficiency. Thus, our theoretical study includes designing data structures and algorithms that can utilize modeling capacities while performing operations as efficiently as possible. 

\subsection{Proof of Theorem~\ref{lemma:sqrt_n_lower_bound}}
We would like to bound $\mathds{E}_{\mD\sim\chi}[\normx{\hat{f}-f_{\mD^j}}]$. Note that both $\hat{f}$ and $f_{\mD^j}$ are random variable (since $\hat{f}$ depends on $f_{\mD^i}$). First, consider $$f_{\mD^j}(q)=\frac{1}{j} \sum_{k\in [j]}I_{\mD_k\in q}=\frac{i}{j}f_{\mD^i}(q) + \frac{j-i}{j}f_{\mD^{i:j}}(q),$$ So that the error is \begin{align*}
    &\mathds{E}_{\mD\sim\chi}[\normx{\hat{f}-\frac{i}{j}f_{\mD^i}(q) + \frac{j-i}{j}f_{\mD^{i:j}}(q)}]\\&=\mathds{E}_{\mD^i\sim\chi}[\mathds{E}_{\mD^{i:j}\sim\chi}[\normx{\hat{f}-\frac{i}{j}f_{\mD^i}(q) - \frac{j-i}{j}f_{\mD^{i:j}}(q)}|\mD^i]].
\end{align*}

Now consider $\mathds{E}_{\mD^{i:j}\sim\chi}[\normx{\hat{f}-\frac{i}{j}f_{\mD^i}(q) - \frac{j-i}{j}f_{\mD^{i:j}}(q)}|\mD^i]$. Given $\mD^i$, $\hat{f}-\frac{i}{j}f_{\mD^i}(q)$ is a fixed quantity. Furthermore, recall that $\arg\min_{c}\mathds{E}[|X-c|]=\text{Med}(X)$ for any random variable $X$, where $\text{Med}(X)$ is a median of $X$ \cite{wasan1970parametric}. Therefore, for any query, 
\begin{align*}
    \mathds{E}_{\mD^{i:j}\sim\chi}&[\normx{\hat{f}-\frac{i}{j}f_{\mD^i}(q) - \frac{j-i}{j}f_{\mD^{i:j}}(q)}|\mD^i]\\&\geq \frac{j-i}{j}\mathds{E}_{\mD^{i:j}\sim\chi}[\text{Med}(f_{\mD^{i:j}}(q))-f_{\mD^{i:j}}(q)].
\end{align*} 

Observe that $f_{\mD^{i:j}}(q)\sim Binomial(j-i, \mathds{P}_{p\sim \chi}(I_{p\in q}))$ and consider any query such that $(j-i)\mathds{P}_{p\sim \chi}(I_{p\in q})$ is an integer, which exists as long as the c.d.f of the distribution is continuous. For such queries, we have $\text{Med}(X)=(j-i)\mathds{P}_{p\sim \chi}(I_{p\in q})$ since mean and median of binomial distributions where $(j-i)\mathds{P}_{p\sim \chi}(I_{p\in q})$ is an integer are equal \cite{janson2021probability}. Let $p_q=\mathds{P}_{p\sim \chi}(I_{p\in q})$. Using the bound on the binomial mean absolute deviation in \citet{berend2013sharp}, we have, when $j-i\geq 2$ and  for any query s.t. $\frac{1}{(j-i)}\leq p_q\leq 1-\frac{1}{j-i}$,
\begin{align*}
    \frac{\sqrt{(j-i)p_q(1-p_q)}}{\sqrt{2}}\leq\mathds{E}_{\mD^{i:j}\sim\chi}[|(j-i)p_q-f_{\mD^{i:j}}(q)|].
\end{align*}
Moreover, setting $p_q=\frac{\floorx{\frac{j-i}{2}}}{j-i}$, we have $$\frac{\sqrt{(j-i)p_q(1-p_q)}}{\sqrt{2}}\geq \frac{\sqrt{j-i}}{4}.$$
\qed

\subsection{Proof of Theorem~\ref{thm:generic_oracle_construction} (formally Theorem~\ref{thm:generic_oracle_construction_formal})}
First, we use the algorithm $\mathcal{A}$ (due to optimizability) to construct the algorithm in definition~\ref{def:dist_learnable} as $\hat{f}=\frac{1}{n}\mathfrak{A}(f_{\mD}, \alpha_{\mathcal{F}\rightarrow\mathcal{G}}(\frac{1}{\sqrt{n}}))$ for $f\in \{r, c\}$ given an input dataset, $\mD$, of size $n$. Let $\vartheta=\alpha_{\mathcal{F}\rightarrow\mathcal{G}}(\frac{1}{\epsilon})$.

Note that since $\mathcal{F}$ has enough representation power to represent $\mathcal{G}$, and since by assumption $\{\chi\in\mathfrak{X}, f_\chi\}\subseteq\mathcal{G}$, we have that, for any $\chi$ there exists $\hat{f}_\chi\in \mathcal{F}_{\vartheta}$ s.t. $\normx{\hat{f}_\chi-f_\chi}_\infty\leq \frac{1}{\sqrt{n}}$. Furthermore, since we approximately optimally find $\hat{f}$, we have $|\hat{f}(x)-\frac{1}{n}f_{\mD}(x)|\leq \frac{1}{\varkappa}|\hat{f}_\chi(x)-\frac{1}{n}f_{\mD}(x)|+\frac{\varkappa'}{\sqrt{n}}$.  Now, to analyze accuracy of $\hat{f}$, observe that, for any input $x$ we have
\begin{align*}
    |\hat{f}(x)-\frac{1}{n}f_{\mD}(x)|&\leq \frac{1}{\varkappa}|\hat{f}_\chi(x)-\frac{1}{n}f_{\mD}(x)| +\frac{\varkappa'}{\sqrt{n}}\\
    &\leq \frac{1}{\varkappa}|\hat{f}_\chi(x)-f_\chi(x)|+\\&\hspace{1.6cm}\frac{1}{\varkappa}|f_\chi(x)-\frac{1}{n}f_{\mD}(x)|+\frac{\varkappa'}{\sqrt{n}}\\
    &\leq \frac{1}{\varkappa\sqrt{n}}+\frac{1}{\varkappa}|f_\chi(x)-\frac{1}{n}f_{\mD}(x)|+\frac{\varkappa'}{\sqrt{n}}.
\end{align*}
We also have 
$$|\hat{f}(x)-f_\chi(x)|\leq |\hat{f}(x)-\frac{1}{n}f_{\mD}(x)|+|f_\chi(x)-\frac{1}{n}f_{\mD}(x)|,$$
So that , 
$$
n|\hat{f}(x)-f_\chi(x)|\leq \frac{\sqrt{n}}{\varkappa}+\frac{2}{\varkappa}|nf_\chi(x)-f_{\mD}(x)|+\sqrt{n}\varkappa'.
$$
By Hoeffeding's inequality, we have
\begin{align}
    \mathds{P}(|nf_\chi(x)-f_{\mD}(x)|\geq \epsilon')\leq e^{-2(\frac{\epsilon'}{\sqrt{n}})^2},
\end{align}
So that 
\begin{align}
    \mathds{P}(n|f_\chi(x)-\hat{f}(x)|\geq \frac{2}{\varkappa}\epsilon'+\sqrt{n}(\frac{1}{\varkappa}+\varkappa'))\leq e^{-2(\frac{\epsilon'}{\sqrt{n}})^2},
\end{align}
And therefore, for some universal constant $\varkappa_2$ and $\epsilon=\Omega(\sqrt{n})$,
 \begin{align}
    \mathds{P}(n|f_\chi(x)-\hat{f}(x)|\geq \epsilon)\leq e^{-\varkappa_2(\frac{\epsilon}{\sqrt{n}}-1)^2}.
\end{align}


\subsection{Proof of Lemma~\ref{lemma:specific_oracle_construction}}
For each distribution class, we show optimizability and representation power of some function class $\mathcal{F}$ that can be used to model the distribution class, which combined with Theorem~\ref{thm:generic_oracle_construction_formal} shows the desired result for both Lemmas~\ref{lemma:specific_oracle_construction} and \ref{lemma:specific_oracle_construction_ce}. Then, for each class, we discuss modeling complexities.

\textbf{Distribution learnability for $\mathfrak{X}_\rho$}. Let $\mathcal{F}$ be the class of piecewise constant functions with uniformly spaced pieces and let $\mathcal{G}$ be the class of real-valued differentiable functions $[0, 1]^d\rightarrow\mathbb{R}$ with gradient bounded by $\rho$. Consider the number of pieces to use for approximation as a hyperparameter. 

\textit{Optimizability}. Given the number of pieces, the function that creates the minimum infinity norm is to place a constant at the mid-point of maximum and minimum values in each interval.  That is, for an interval $I\subseteq [0, 1]^d$, the constant approximating $g$ over with the lowest infinity norm $I$ is $\frac{1}{2}(\min_{x\in I}g(x)+\max_{x\in I}f(x)))$. Note that this function has error at most $\max_{x\in I}g(x)-\frac{1}{2}(\min_{x\in I}g(x)+\max_{x\in I}f(x)))=\frac{1}{2}(\max_{x\in I}f(x)-\min_{x\in I}g(x))$. For efficiency purposes, instead of the optimal solution, we let the constant for the piece responsible for $I$ be $g(p)$ for some $p\in I$. Note that for all $x\in I$ $|g(p)-g(x)|\leq |\max_{x\in I}g(x)-\min_{x\in I}g(x)|$, so that this construction gives us a $\frac{1}{2}$-approximation of the optimal solution. 
 
\textit{Representation Power}.  Define $\alpha_{\mathcal{F}\rightarrow\mathcal{G}}(\epsilon)=\frac{\sqrt{d}\rho}{\epsilon}$. We show that for any $g\in \mathcal{G}$ and any $\epsilon>0$, there is a function $\hat{f}\in \mathcal{F}_{\alpha(\epsilon)}$ s.t. $\normx{\hat{f}-g}_\infty\leq \epsilon$. This function is the optimal solution as constructed above. To see why the error is at most $\epsilon$, consider a partition over $I$ with $j$-th dimension $[p_{j, i}, p_{j, i+1}]$, where $p_{j,i+1}-p_{j, i}=\frac{\epsilon}{\rho}$, and let $x_1$ and $x_2$ be the two points in $I$ that, respectively, achieve the minimum and maximum of $g$ in $I$. For any point in $x\in I$, our function approximator answer $\hat{f}(x)=\frac{1}{2}(g(x_1)+g(x_2))$. We have that 
\begin{align*}
    |\hat{f}(x)-g(x)|&=|\frac{1}{2}(g(x_1)+g(x_2))-g(x)|\\
    &\leq \max\{g(x_2)-g(x), g(x)-g(x_1)\}\\
    &\leq \normx{g'(x)}_2\normx{x_2-x}_2\\
    &\leq \rho(\sqrt{d}\frac{\epsilon}{\sqrt{d}\rho})=\epsilon.
\end{align*}

\textit{Model Complexity}. The inference time, $\mathcal{T}_n^{\mathfrak{X}}$, is constant independent of the number of pieces used. The space complexity is the number of pieces multiplied by the space to store each constant. Given that  $g$ consists of integers between 0 to $n$, $\mathcal{S}_n^{\mathfrak{X}}$ can be stored in $O(\sqrt{d}(\rho\sqrt{n})^d\log n)$. Finally, for rank operation, the algorithm that outputs the function optimizer makes $\rho\sqrt{n}$ calls to $g$, so that building the function approximator can be done in $O(\rho\sqrt{n}\log n)$, assuming the data is sorted (so that each call to $g$ takes $O(\log n)$). For cardinality estimation there are $O(\sqrt{d}(\rho\sqrt{n})^d\log n)$ calls to the cardinality function, where each call in the worst case takes $O(n)$ (this can optimized by building high dimensional indexes). Thus, in this case, $\mathcal{B}_n^{\mathfrak{X}}=O(\sqrt{d}(\rho\sqrt{n})^d\log n)$

\textbf{Distribution learnability for $\mathfrak{X}_{l}$}. Let $\mathcal{F}$ and $\mathcal{G}$ be the class of piecewise linear functions with at most $l$ pieces (not necessarily uniformly spaced pieces). Trivially, $\mathcal{F}$ has enough representation power to represent $\mathcal{G}$, thus, it remains to show optimizability and model complexity.

\textit{Optimizability}. The PLA algorithm, $\mathbf{P}(\epsilon)$ by~\cite{orourke1981line}, used in PGM index~\cite{ferragina2020pgm}, is able to find the piecewise linear solution with the smallest number of pieces given an error $\epsilon$. Here, we want to achieve the opposite, i.e., given a number of pieces find piecewise linear approximation with smallest error. Note that $\epsilon$ is in the range $0$ to $1$, and we can do a binary search on the values of $\epsilon$, for each calling $\mathbf{P}(\epsilon)$ until we find the smallest $\epsilon$ where $|\mathbf{P}(\epsilon)|\leq l$. Note that since suboptimality of $O(\frac{1}{\sqrt{n}})$ in $\epsilon$ is allowed, wee can discretize $[0, 1]$ to $\sqrt{n}$ groups, and only do binary search over this discrete set, which takes $O(\log(\sqrt{n}))$ calls to $\mathbf{P}(\epsilon)$, and each call takes $O(n)$ operations~\cite{ferragina2020pgm} on a sorted array, so that $\mathcal{F}$ is optimizable to with the algorithm running in $O(n\log n)$ 
 
\textit{Model Complexity}. The learning time $n\mathcal{B}_n^{\mathfrak{X}}=O(n\log n)$ is discussed above. The algorithm always returns $l$ pieces which can be evaluated in $\mathcal{T}_n^{\mathfrak{X}}=O(\log l)$ time. Note the each linear piece can be adjusted to cover an interval starting and ending at points in the dataset (so the interval can be stored as pointers to corresponding dataset item). Moreover, the beginning and end of each line can be adjusted to be an integer (since the rank function only returns integers), similar to~\cite{ferragina2020pgm}, so that the lines can be stored in $\mathcal{S}_n^{\mathfrak{X}}=O(l\log n)$. 

\textbf{Distribution Learnability for $\mathfrak{X}_c$}. Trivially, the class $\mathcal{F}$ containing the distribution operation function has enough approximation power for $\mathfrak{X}_c$ and is optimizable with $\mathcal{S}_n^{\mathfrak{X}}, \mathcal{T}_n^{\mathfrak{X}}, \mathcal{B}_n^{\mathfrak{X}}$ all $O(1)$.

\subsection{Proofs for Learned Indexing}
\input{proof_indexing}

\subsection{Proofs for Cardinality Estimation}
\input{proof_ce}

\subsection{Sorting}
\input{proof_sorting}

%% file: proof_indexing.tex
\begin{algorithm}[t]
\begin{algorithmic}[1]
\Require New element to be inserted in tree rooted at $N$
\Ensure Balanced data structure
\Procedure{Query}{$q$, $N$}
    \If{$N$ is a leaf node}
        \State \Return \textsc{BinarySearch}($N.\texttt{content}$)
    \EndIf
    \State $\hat{i}\leftarrow N.\hat{f}(p)$
    \State $i\leftarrow \textsc{ExpSearch}(p, \hat{i}, N.\texttt{content})$
    \State $j\leftarrow \textsc{BinarySearch}(p, N.\texttt{children}[i])$
    \State\Return \textsc{Query}($q$, $N.\texttt{children}[i][j]$)
\EndProcedure
\caption{Dynamic Learned Index Query}\label{alg:dynamic_learned_index_query}
\end{algorithmic}
\end{algorithm}
\begin{algorithm}[t]
\begin{algorithmic}[1]
\Require New element, $p$ to be inserted in tree rooted at $N$
\Ensure Index with $p$ inserted
\Procedure{Insert}{$p$, $N$}
    \State $N.\texttt{counter}$++
    \If{$N.\texttt{children}$ \textbf{is} \texttt{NULL}}
        \State \textsc{InsertContent}($p$, $N.\texttt{content}$)
        \State\Return
    \EndIf
    \State $\hat{i}\leftarrow N.\hat{f}(p)$
    \State $i\leftarrow \textsc{ExpSearch}(p, \hat{i}, N.\texttt{content})$
    \State $j\leftarrow \textsc{BinarySearch}(p, N.\texttt{children}[i])$
    \State \textsc{Insert}($p$, $N.\texttt{children}[i][j]$)
    \If{$N.\texttt{counter}=N.\texttt{max\_points}$}
        \State $A\leftarrow$ the sorted array in the index rooted at $N$
        \If{$N$ has no parent}
            \State\Return $\textsc{Rebuild}(A)$
        \EndIf
        \State $P\leftarrow$ parent of $N$
        \State $i_p\leftarrow$ index of $N$ in $P.\texttt{children}$
        \State Remove $N$ from $P.\texttt{children}[i_p]$
        \State $N_1 \leftarrow$ \textsc{Rebuild}($A[:N.\texttt{max\_points}/2]$)
        \State $N_2\leftarrow$ \textsc{Rebuild}($A[N.\texttt{max\_points}/2:]$)
        \State Insert $N_1$ and $N_2$ in $P.\texttt{children}[i_p]$
        
    \EndIf 
\EndProcedure
\caption{Dynamic Learned Index Insertions}\label{alg:dynamic_learned_index_insert}
\end{algorithmic}
\end{algorithm}

\begin{algorithm}[t]
\begin{algorithmic}[1]
\Require A sorted array, $A$
\Ensure A learned index rooted at a new node $N$
\Procedure{Rebuild}{$A$}
    \State $N\leftarrow$ new node
    \State $k\leftarrow |A|$
    \State $N.\texttt{counter}\leftarrow k$
    \If{$k\leq \varkappa$}
        \State $N.\texttt{content}\leftarrow A$
        \State \Return $N$
    \EndIf    
    \State $N.\texttt{max\_points}\leftarrow2k$
    \State $N.\hat{f}=\mathcal{A}(A)$
    \State $N.\texttt{content}\leftarrow A[::\sqrt{k}]$
    \For{$i$ in $\sqrt{k}$}
        \State $N_c\leftarrow \textsc{Rebuild}(A[i k:(i+1)k])$
        \State $N.\texttt{children}.append(N_c)$
    \EndFor
    \State\Return $N$
\EndProcedure
\caption{Procedure for for Rebuilding Root}\label{alg:dynamic_learned_index_rebuild}
\end{algorithmic}
\end{algorithm}

\begin{figure}[t]
    \centering
    \includegraphics[width=\columnwidth]{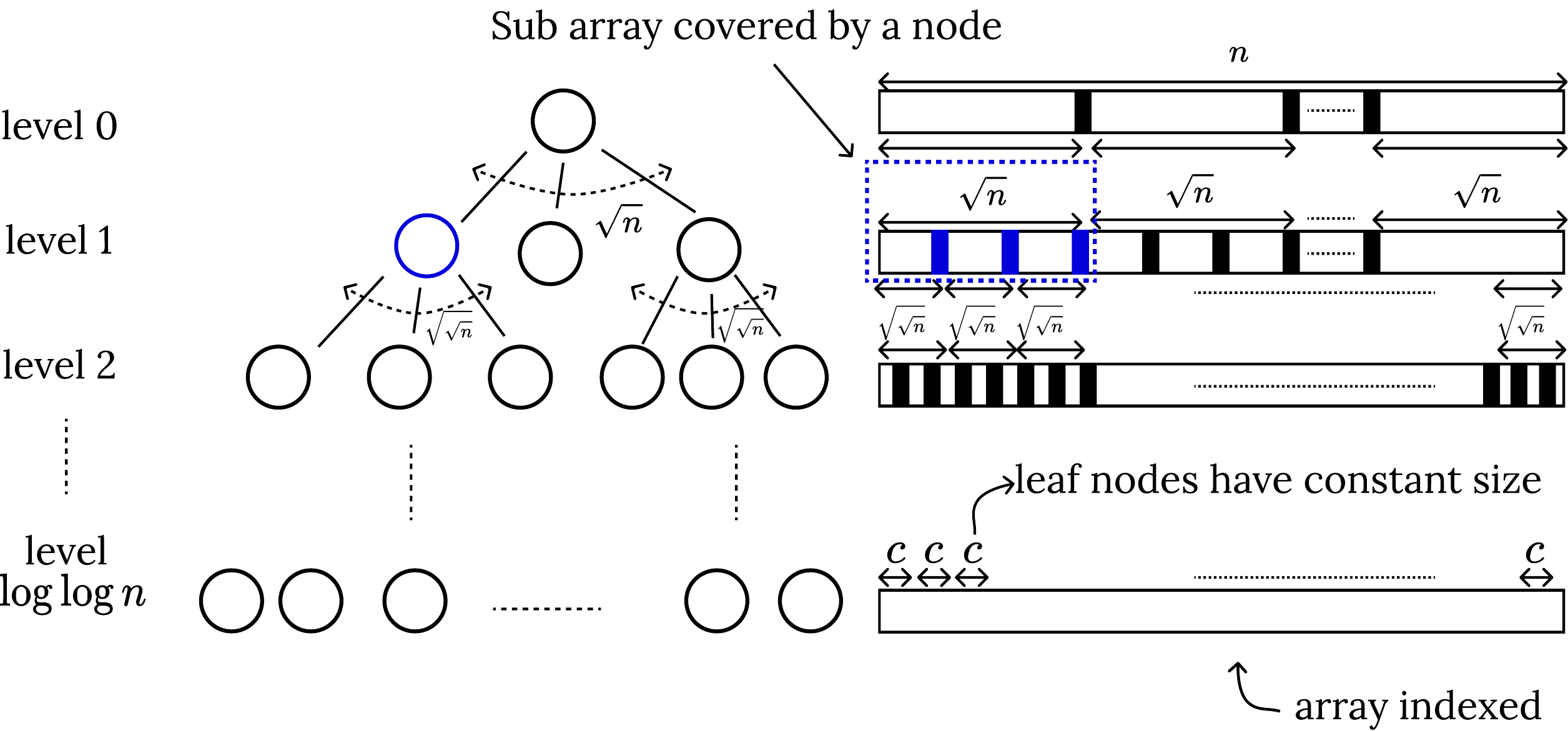}
    \vspace{-0.4cm}
    \caption{Structure of the learned dynamic index}
    \label{fig:index}
\end{figure}
\begin{figure}[t]
    \centering
    \includegraphics[width=\columnwidth]{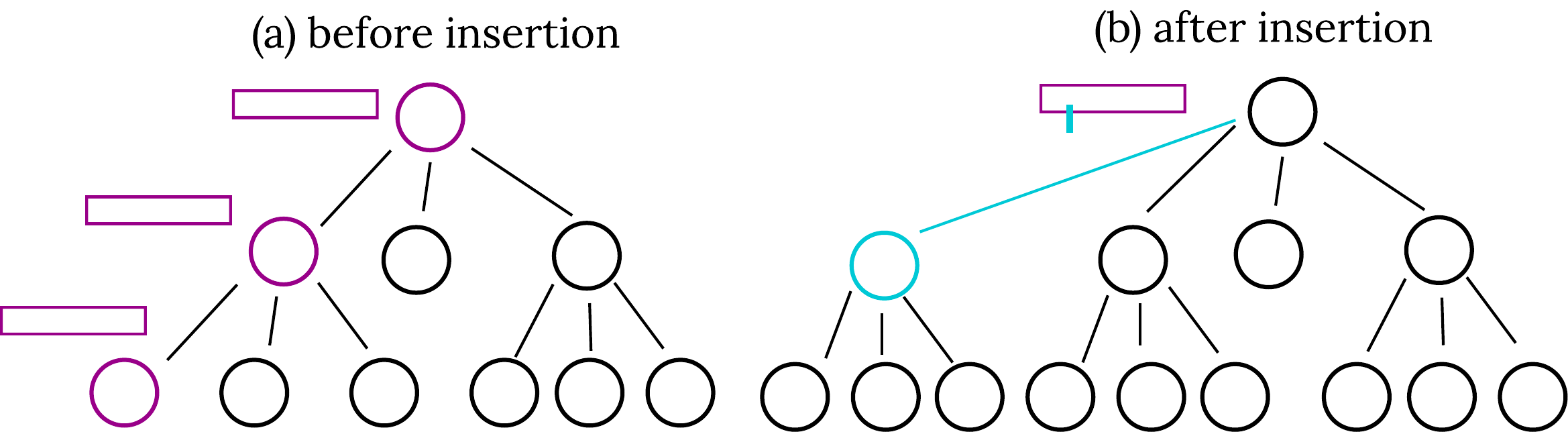}
    \vspace{-0.4cm}
    \caption{Insertion Causing a split in index}
    \label{fig:index_split}
\end{figure}

\subsubsection{Index Operations}
The index supports two operations, \textsc{Query} and \textsc{Insert}, which are presented in Algs.~\ref{alg:dynamic_learned_index_query} and~\ref{alg:dynamic_learned_index_insert}. Recall that $\mathcal{A}$ is an algorithm defined in Definition~\ref{def:dist_learnable} and exists due to distribution learnability of $\mathfrak{X}$. The index builds a tree structure similar to \cite{ding2020alex}, with each node containing a model, and a set of children. An overview of the tree architecture is shown in Fig.~\ref{fig:index}. Each node can be seen to cover a subarray of the original indexed array. If the size of the covered subarray is $k$ elements, then the node will have $\sqrt{k}$ children, where the subarray of size $k$ is equally divided between the children (so each child covers $\sqrt{k}$ elements). The root node covers the entire $n$ elements of the array, and therefore has $\sqrt{n}$ children. As can be seen, the number of children of the nodes decreases as we go down the tree. A node won't have any children if its covered subarray is smaller than some constant $c$. Moreover, shown as black elements in the figure, each parent node stores the minimum value of the subarray covered by each of its children in an array called the node's $\texttt{content}$. Thus, the node's $\texttt{content}$ can be used to traverse the tree. When a node's model predicts which child the node should travel, the node first checks with its $\texttt{content}$ to make sure it is the correct node. This is done by doing an exponential search on the node's $\texttt{content}$. 

During insertions, each node keeps a counter of the number of points inserted through it. If a node is at level $i$, then at most $k^{2^{-i}}$ elements are allowed in the node for $i>0$, where $k$ is the size of the dataset at the time of construction of the current root node (root node's are periodically rebuilt). If the number of insertions reaches $k^{2^{-i}}$, the node splits. When a node splits the subarray it covers is split into two, and an entirely new subtree is built for each half of the subarray, rebuilding all models. To avoid splits affecting parent nodes, as Fig.~\ref{fig:index_split} shows, the newly created node is appended to the list of children (we'll discuss how exactly this is done later). Finally, the root is rebuilt every time its size doubles. 

To support the splitting discussed above, the children are arranged in a two dimensional array $N.\texttt{children}$. We refer to children pointed to in $N.\texttt{children}[i]$ as children in the $i$-th child slot. Each child slot contains a sorted list of at least one, but a variable number, of children, where the list is kept sorted using binary trees. To find which node to traverse, we first find the correct node slot with the help of the learned model, and then use the binary tree in the node slot to find the correct child. Leaf nodes have $\texttt{content}$ storing the data. The index keeps a counter at each node, and periodically rebuilds the tree rooted at a node.  Thus, the two operations are performed as follows.

\textbf{Query}. Performing queries on an index with root node $N$ is similar to performing queries with a B-tree, where nodes are recursively traversed until reaching a leaf node. The only difference is how we decide which child to search. This is done by, for each non-leaf node, first asking a model to estimate which child to search. Then, the model estimate is corrected by performing a local exponential search. Since a child might have been split, the search then uses the binary tree at the correct node slot to find the correct child.

\textbf{Insertions}. Insertions first traverses the tree similar to the queries, with the extra addition that a counter in each node is incremented if a new element is inserted in that node. If the counter of a node passes the maximum size of the node, the node is split into two, and the parent meta data for the corresponding node slot is updated. The only exception is the root node, which does not split, but triggers a full rebuild of the entire tree.

\subsubsection{Query Time}
Queries are performed by recursively searching each node, where a single node per level is queried. Consider the number of operations performed at the $i$-th level. Each level performs a model inference, exponential search on the node's content and a binary search on the node's extension. We consider each separately.

\textbf{Model Inference}. A model at the $i$-th level is built on at most $n^{\frac{1}{2^i}}$ elements. Thus, the inference time is $O(\mathcal{T}_{{n^{2^{\shortminus i}}}}^{\mathfrak{X}})$.

\textbf{Exponential Search}. The time complexity of the exponential search step depends on the accuracy of the model estimate. We show that, on expectation, the time complexity is constant.

Note that neither the content of each node, nor its model gets modified by insertions, unless a node is rebuilt. Thus, we only need to show the statement for right after the node construction. Assume the node is constructed for some array $A$ with $|A|=k$ for some integer $k$. 

Assume the $i$-th element of $A$ is a sample originally obtained from the distribution $\chi_i$, let $R=[l, u]$ define the range the elements $A$ have fallen into passed to the algorithm. 
The elements of $A$ are independent samples from conditional distributions $\chi_1\mid R, ..., \chi_k\mid R$, respectively. Let $\chi_R=\{\chi_1\mid R, ..., \chi_k\mid R\}$.

Applying the extension of the DKW bound to independent but non-identical random variables \citep[Chapter~25.1]{shorack1986empirical} we have that 
\begin{align}\label{ineq:indexing:dkw}
\mathds{P}(\normx{kr_{\chi_R}-r_{A}}_\infty\geq \sqrt{\frac{k}{2}}\epsilon)\leq 2e^{-\epsilon^2+1}.
\end{align}
Furthermore, for $\hat{r}$, the model of $r_{\chi_R}=\frac{1}{k}\sum_{i=1}^kr_{\chi_i\mid R}$, obtained from $\mathcal{A}$ and by the accuracy requirement of distribution learnability, we have for any $\epsilon\geq\sqrt{\varkappa_2}(\frac{1}{\sqrt{k}}-1)$
\begin{align}\label{ineq:indexing:oracle}
\mathds{P}(\normx{kr_{\chi_R}-\hat{r}}_\infty\geq \sqrt{k}(\frac{\epsilon}{\sqrt{\varkappa_2}}+1))\leq \varkappa_1e^{-\epsilon^2}.    
\end{align}
By union bound on Ineq.~\ref{ineq:indexing:dkw} and ~\ref{ineq:indexing:oracle}, and the triangle inequality, for any $\epsilon\geq\sqrt{k}$, we have
\begin{align}
    \mathds{P}(\normx{\hat{r}-r_{A}}_\infty\geq \epsilon)\leq \varkappa_1'e^{-\varkappa'_2(\frac{\epsilon}{\sqrt{k}}-1)^2},
\end{align}
where $\varkappa'_1=2e+\varkappa_1$ and $\varkappa'_2=\frac{2\varkappa_2}{(\sqrt{2}+\sqrt{\varkappa_2})^2}$.

Now let $N(q)$ be the number of operations by performed by exponential search for a query $q$. Recall that the content of the node is $A_c=A[::\sqrt{k}]$ and that we use $\hat{r}_{A_c}=\ceilx{\frac{1}{\sqrt{k}}\hat{r}(q)}$ as the start location to start searching for $q$ in $A_c$ with exponential search. We have the true location of $q$ in $A_c$ is $r_{A_c}(q)=\ceilx{\frac{1}{\sqrt{k}}r_A(q)}$. 
Observe that, for any $q$
\begin{align}\label{ineq:error_due_to_rounding}
    \normx{r_A-\hat{r}}_\infty> (\hat{r}_{A_c}(q)-r_{A_c}(q)|-2)\sqrt{k},
\end{align}
and that it is easy to see that for exponential search we have
\begin{align}\label{ineq:exp_search_error_ops}
|\hat{r}_{A_c}(q)-r_{A_c}(q)|\geq 2^{\frac{N(q)}{2}-1},
\end{align}
Combining which we get, for any query $q$,
\begin{align*}
    \normx{r_A-\hat{r}}_\infty> (2^{\frac{N(q)}{2}-1}-2)\sqrt{k}.
\end{align*}
So that, for any $i$, $N(q)\geq i$ implies that $\normx{r_A-\hat{r}}_\infty> (|2^{\frac{i}{2}-1}-2)\sqrt{k}$
Thus, we have
\begin{align*}
    \mathds{E}_{A\sim\chi}[N(q)] &= \sum_{i=1}^{2\log k}\mathds{P}_{A\sim\chi}(N(q)\geq i)\\
    &\leq  \sum_{i=1}^{2\log k}\mathds{P}_{A\sim\chi}(\normx{r_A-\hat{r}}_\infty> (|2^{\frac{i}{2}-1}-2)\sqrt{k})\\
    &\leq 5+  \varkappa'_1\sum_{i=6}^{2\log k}e^{-\varkappa'_2(2^{\frac{i}{2}-1}-3)^2}=O(1)\\
\end{align*}

Thus, the expected time searching performing exponential search is $O(1)$.

\textbf{Searching Node's Extension}. Next, we study the expected time for searching the additional list added to the nodes. Let $L$ be the size of the list. Note that the lists are created for all the nodes except the root. Moreover, a non-root node at level $i$ has capacity $cn^{\frac{1}{2^i}}$, it will split every $cn^{\frac{1}{2^i}}$ insertions into the parent, with each split adding an element to the parent's extension list. Furthermore, the node gets rebuilt after every $k=n^{\frac{1}{2^{i-1}}}$ insertions into its parent. Thus, if $k_N$ elements out of $n^{\frac{1}{2^{i-1}}}$ get inserted into a node slot $N$, the number of splits for that slot will be $\floorx{\frac{k_N}{cn^{\frac{1}{2^i}}}}$. 

Assume the $k$ new insertions into the parent, $N_p$ of $N$ since the last rebuild of the parent were from r.v.s with distribution $\chi_1, ..., \chi_k$. Given that they fall in $N_p$, their conditional distribution is $\chi_1\mid R, ..., \chi_k\mid R$ for $R$ defining an interval for which the node $N_P$ was built. Let $\chi'_R=\{\chi_1\mid R, ..., \chi_k\mid R\}$. Furthermore, let $\chi_R$ be the original distribution the model of in the parent node was created based on. Let $A'$ be the set of $k$ insertions and let $A$ be the set of points based on which the parent of $N$ was built. The number of insertions out of the $k$ new insertions into the $j$-th node slot is $r_{A'}(N_j)-r_{A'}(N_{j-1})$. To study this quantity, recall that the $j$-th slot was created so that 
\begin{align}\label{ineq:extension:rank_creation}
r_A(N_j)-r_A(N_{j-1})=n^{\frac{1}{2^i}}.    
\end{align}
Thus, we first relate $r_A$ and $r_A'$. 


The elements of $A$ and $A'$ are independent so that applying the extension of the DKW bound to independent but non-identical random variables \citep[Chapter~25.1]{shorack1986empirical} for both $A$ and $A'$ we have
\begin{align}
\label{ineq:indexing:extension:dkwA}\mathds{P}(\normx{kr_{\chi'_R}-r_{A'}}_\infty\geq \sqrt{\frac{k}{2}}\epsilon)\leq 2e^{-\epsilon^2+1}, \\
\label{ineq:indexing:extension:dkwA'}\mathds{P}(\normx{kr_{\chi_R}-r_{A}}_\infty\geq \sqrt{\frac{k}{2}}\epsilon)\leq 2e^{-\epsilon^2+1}.
\end{align}
Moreover, assume TV($\chi_i\mid R, \chi_j\mid R$)$\leq\delta$ for all $i, j$, we have that $\normx{r_\chi-r_{\chi'}}_\infty\leq\delta$. Combining this with Ineq.~\ref{ineq:indexing:extension:dkwA} and \ref{ineq:indexing:extension:dkwA'} and using the triangle inequality, we have
\begin{align}\label{ineq:extension:rank_diff}
    \mathds{P}(\normx{r_{A'}-r_{A}}_\infty\geq \sqrt{2k}\epsilon+\delta k)\leq 4e^{-\epsilon^2+1}.
\end{align}
Finally, combining Ineq.~\ref{ineq:extension:rank_diff} with Eq.~\ref{ineq:extension:rank_creation}, and recalling that $k=n^{2^{-i+1}}$ implies that 
\begin{align*}
    \mathds{P}(r_{A'}(N_j)-r_{A'}(N_{j-1})\geq n^{2^{-i}}(1+2\sqrt{2}\epsilon)+2&\delta n^{2^{-i+1}})\\&\leq 4e^{-\epsilon^2+1}.
\end{align*}
And therefore
\begin{align*}
    \mathds{P}(\floorx{\frac{r_{A'}(N_j)-r_{A'}(N_{j-1})}{n^{2^{-i}}}}\geq 2\sqrt{2}\epsilon+2\delta n^{2^{-i}})\leq 4e^{-\epsilon^2+1},
\end{align*}
Where $\floorx{\frac{r_{A'}(N_j)-r_{A'}(N_{j-1})}{n^{2^{-i}}}}$ is the number of splits of the $j$-th node slot. Let $S_j$ denote this random variable. We have
\begin{align*}
    \mathds{E}[S_j]&=\sum_{i=0}^k\mathds{P}(S_j\geq i)\\
    &\leq 2\delta n^{2^{-i}}+\sum_{i=0}^{k-2\delta n^{2^{-i}}}\mathds{P}(S_j\geq i+2\delta n^{2^{-i}})\\
    &\leq 2\delta n^{2^{-i}}+\sum_{i=0}^{k-2\delta n^{2^{-i}}}4e^{-\frac{i^2}{8}+1}=O(\delta n^{2^{-i}})
\end{align*}
Finally, we are interested in $\mathds{E}[\log(S_j)]\leq \log(\mathds{E}[S_j])= O(\log (n^{\frac{1}{2^i}}\delta))$.

\textbf{Total Query Time}. Thus, the expected time to search a node at the $i$-th level to find its children is $O(\mathcal{T}_{{n^{2^{\shortminus i}}}}^{\mathfrak{X}}+\log(n^{-2^i}\delta))$. Thus, the total time to search the tree is $O(\sum_{i=1}^{\log\log n}\mathcal{T}_{{n^{2^{\shortminus i}}}}^{\mathfrak{X}}+\log(n^{-2^i}\delta_i))$. We can bound this as $O(\mathcal{T}_{{n}}^{\mathfrak{X}}\log\log n+\log(n\bar{\delta}))$, where $\bar{\delta}=\min\{\delta, \delta_c^{\log\log n}\}$.

\subsubsection{Insertion Time}
Note that insertion time is equal to query time plus the total cost of rebuilds. Next, we calculate the cost of rebuilds.

Let $T(N)$ be the amortized cost of inserting $N$ elements into a tree that currently has $N$ elements and was just rebuilt at its root, so that $NT(N)$ will be the total insertion cost for the $N$ elements. Note that the amortized cost of all $n$ insertions starting from a tree with one 1 element is at most $\frac{1}{n}\sum_{i=0}^{\log n} \frac{n}{2^i}T(\frac{n}{2^i}) \leq T(n)\sum\frac{1}{2^i}=O(T(n))$. 

Thus, we only need to study $T(n)$. Note that when inserting $n$ elements into a tree that currently has $n$ elements and was just rebuilt at its root, the height of the tree remains constant throughout insertions. Furthermore, at the $i$-th level, $i\geq 0$, there will be at most $\frac{n}{n^{2^{-i}}}$ rebuilds and each rebuild costs $$\sum_{j=i}^{\log\log n}\frac{2n^{2^{-i}}}{n^{2^{-j}}}\mathcal{B}_{2n^{2^{\shortminus j}}}^\mathfrak{X}.$$ Thus, we have the amortized cost of all rebuilds is 
\begin{align*}
    \frac{1}{n}\sum_{i=0}^{\log\log n}\frac{n}{n^{2^{-i}}}\sum_{j=i}^{\log\log n}\frac{2n^{2^{-i}}}{n^{2^{-j}}}\mathcal{B}_{2n^{2^{\shortminus j}}}^\mathfrak{X}=\\O(\sum_{i=0}^{\log\log n}(i+1)\frac{\mathcal{B}_{2n^{2^{\shortminus i}}}^\mathfrak{X}}{n^{2^{-i}}}).
\end{align*}
Thus, the total cost of insertions is $O(\mathcal{T}_n^\mathfrak{X}\log\log n+\log(\bar{\delta} n)+\frac{\mathcal{B}_n^\mathfrak{X}}{n}\log^2\log n))$. 
\qed

\subsubsection{Space Overhead}
After $n$ insertions, we will have $\log\log n$ levels. Right after the root was rebuilt, level $i$ has at most $\frac{n}{n^{2^{-i}}}$ models. If $n$ further insertions are performed, each level will have at most $\frac{n}{n^{2^{-i}}}$ new models. Thus, the total size of the models at level $i$ is at most $2\frac{n}{n^{2^{-i}}}\mathcal{S}_{n^{2^{\shortminus i}}}^\mathfrak{X}$, and thus the total size of all models is $O(n\sum_{i=0}^{\log\log n}\frac{\mathcal{S}_{n^{2^{\shortminus i}}}^\mathfrak{X}}{n^{2^{-i}}})$. Furthermore, the total number of nodes in the tree is $O(n)$, and for each node we store a pointer to it and its lower and upper bounds, as well as a counter, which can be done in $O(\log n)$. Thus, the total space consumption is $O(n(\log n+\sum_{i=0}^{\log\log n}\frac{\mathcal{S}_{n^{2^{\shortminus i}}}^\mathfrak{X}}{n^{2^{-i}}}))$.

\subsection{Proof of Corollary~\ref{corollary:indexing_specific_construction}}
Observe that for any distribution in $\mathfrak{X}$, we have that $\bar{\chi|R}$ for any interval $R$ has p.d.f at most $\frac{\rho_2}{\rho_1}$.  According to Lemma~\ref{lemma:specific_oracle_construction}, distributions with p.d.f at most $\frac{\rho_2}{\rho_1}$ are distribution learnable. Substituting the complexities proves Corollary.

%% file: proof_ce.tex
\subsubsection{High Dimensions (Theorem~\ref{thm:dynamic_ce_hd_oracle})}
\textbf{Construction}. By distribution learnablility we have an algorithm $\mathcal{A}$ that builds a model that we use for estimation. To prove the lemma, we use $\mathcal{A}$ and periodically rebuild models to answer queries. Specifically, $\mathcal{A}$ is called  every $k$ insertions, where if $\delta\geq \frac{2\varkappa}{\sqrt{n}}$, $k=\frac{\phi-\varkappa}{2\varkappa\delta}\sqrt{n}$, and when $\delta\leq \frac{2\varkappa}{\sqrt{n}}$, $k=n\times\min\{(\frac{\phi-\varkappa}{\varkappa(1+2\varkappa)})^2, 1\}$. That is, if currently there are $n$ points inserted, and we insert $n'$ new points, for $n'<k$, the algorithm will answer queries as $(n+n')\hat{c}$. However, when $n'=k$, the algorithm rebuilds the model and starts answering queries using the new model. 

\textbf{Query Time and Space Consumption}. We use a single model with no additional data structure, so query time is $O(\mathcal{T}_{n}^\mathfrak{X})$ and space complexity is $O(\mathcal{S}_{n}^\mathfrak{X})$.
 
\textbf{Insertion Complexity}. To analyze the cost of insertions, first consider, $T(n)$, the total number of operations when we insert $n$ new elements in a data structure that already has $n$ elements. Consider the two cases where $\delta\geq \frac{2\varkappa}{\sqrt{n}}$ and $\delta\leq \frac{2\varkappa}{\sqrt{2}}$. In the first case, we have the we rebuild the model every $\frac{\phi-\varkappa}{2\varkappa\delta}\sqrt{n}$ insertions, so that there are at most $\frac{n}{\frac{\phi-\varkappa}{2\varkappa\delta}\sqrt{n}}=\frac{2\varkappa\delta}{\phi-\varkappa}\sqrt{n}$ rebuilds. If $\delta\geq \frac{2\varkappa}{\sqrt{2}}$, we rebuild the model every $\rho n$ times, so that the total number of rebuilds is $\frac{1}{\rho}$. Ensuring that $\phi\geq\varkappa+1$, we have that $\frac{1}{\rho}\leq \varkappa^2(2\varkappa+1)^2$. In either case, each rebuild costs $O(\mathcal{B}_{2n}^\mathfrak{X})$ and besides rebuilds insertions takes constant time. Thus, if $\delta\geq \frac{2\varkappa}{\sqrt{n}}$, $T(n)=O(n+\frac{\delta}{\phi}\sqrt{n}\mathcal{B}_{2n}^\mathfrak{X})$ and if $\delta\leq \frac{2\varkappa}{\sqrt{2}}$ we have $T(n)=O(n+\mathcal{B}_{2n}^\mathfrak{X})$.


Next, to analyze the total runtime of starting from 0 elements and inserting $n$ new elements, we have that the amortized insertion is $\frac{1}{n}\sum_{i=1}^{\log n}T(\frac{n}{2^i})$. Now if $\delta\geq \frac{2\varkappa}{\sqrt{n}}$, this is $O(\frac{1}{n}\sum_{i=1}^{\log n}\frac{n}{2^i}+\frac{\delta}{\phi}\sqrt{\frac{n}{2^i}}\mathcal{B}_{\frac{2n}{2^i}}^\mathfrak{X})=O(\frac{1}{n}\sum_{i=1}^{\log n}\frac{\delta}{\phi\sqrt{n2^i}}\mathcal{B}_{\frac{2n}{2^i}}^\mathfrak{X})=O(n+\frac{\delta}{\phi}\sqrt{n}\mathcal{B}_{2n}^\mathfrak{X})$. Furthermore, if $\delta\leq \frac{2\varkappa}{\sqrt{n}}$, this is $O(n+\mathcal{B}_{2n}^\mathfrak{X})$. Thus, the amortized insertion cost is $O(\max\{\frac{\delta}{\phi\sqrt{n}}, \frac{1}{n}\}\mathcal{B}_{n}^\mathfrak{X})$

\textbf{Accuracy}. We show that if $\delta\geq \frac{2\varkappa}{\sqrt{n}}$, rebuilding the model every $\frac{\phi-\varkappa}{2\varkappa\delta}\sqrt{n}$, and when $\delta\leq \frac{2\varkappa}{\sqrt{n}}$ rebuilding the model every $\rho n$ for $\rho=\min\{(\frac{\phi-\varkappa}{\varkappa(1+2\varkappa)})^2, 1\}$ insertions is sufficient to answer queries with error at most $\phi\sqrt{n}$, whenever $\phi\geq \varkappa+1$. 

Assume a model was built using dataset $\mD^{i}$, i.e., after $i$ insertions. We study the error in answering after $k$ new insertions, so that the goal is to answer queries on $\mD^j$, $j=i+k$. Let $\chi$ and $\chi'$ be the distributions so that $\mD^i\sim\chi$ and $\mD^{i:j}\sim\chi'$.

Consider a model, $\hat{c}$ that was built on $\mD^i$ using $\mathcal{A}$,  so we have 
\begin{align}\label{label:ce_hd_oracle}
\mathds{P}(i\normx{\hat{c}-c_\chi}_\infty\geq \sqrt{i}(\frac{\epsilon}{\sqrt{\varkappa_2}}+1))\leq \varkappa_1e^{-\epsilon^2}
\end{align}
We are interested 
\begin{align}
\nonumber    \normx{j\hat{c}-c
_{\mD^j}}&=\normx{i\hat{c}+k\hat{c}-c_{\mD^i}-c_{\mD^{i:j}}}\\&\leq\normx{i\hat{c}-c_{\mD^i}}+\normx{k\hat{c}-c_{\mD^{i:j}}}.\label{ineq:ce_hd_decompose}
\end{align} For the first term, by Hoeffding's inequality we have 
\begin{align}
    \mathds{P}(|ic_{\chi}(q)-c_{\mD^i}(q)|\geq\sqrt{i}\epsilon)\leq e^{-2\epsilon^2},
\end{align}
Which combined with Ineq.~\ref{label:ce_hd_oracle} gives
\begin{align}\label{ineq:ce_hd_modeling_error}
    &\mathds{P}\Big(|i\hat{c}(q)-c
_{\mD^i}(q)|\geq \sqrt{i}((1+\frac{1}{\sqrt{\varkappa_2}})\epsilon+1)\Big)\\
    &\hspace{5cm}\nonumber\leq (1+\varkappa_1)e^{-2\epsilon^2}.
\end{align}
For the second term, again by Hoeffding's inequality we have
\begin{align}\label{label:ce_hd_hf}
    \mathds{P}(|kc_{\chi}(q)-c_{\mD^{i:j}}(q)|\geq\sqrt{k}\epsilon)\leq e^{-2\epsilon^2},
\end{align}
We also have that $\normx{\chi-\chi'}\leq \delta$, which combined with Ineq.~\ref{label:ce_hd_hf} gives
$$\mathds{P}(|k c_{\chi}(q)-c_{\mD^{i:j}}(q)|\geq\sqrt{k}\epsilon+k\delta)\leq e^{-2\epsilon^2},$$
And therefore, using Ineq.~\ref{label:ce_hd_oracle}, we have
\begin{align}\label{ineq:ce_hd_generalization_error}
    &\mathds{P}\Big(|k\hat{c}(q)-c_{\mD{i:j}}(q)|\geq \sqrt{k}\epsilon+\frac{k}{\sqrt{i}}(\frac{\epsilon}{\sqrt{\varkappa_2}}+1)+k\delta\Big)\nonumber\\
    &\hspace{4cm}\leq (\varkappa_1+1)e^{-2\epsilon^2}.
\end{align}
Combining Ineq.~\ref{ineq:ce_hd_decompose}, \ref{ineq:ce_hd_modeling_error} and \ref{ineq:ce_hd_generalization_error}, we have 
\begin{align*}
    &\mathds{P}\Big(|j\hat{c}(q)-c_{\mD{j}}(q)|\geq (\sqrt{i}+\frac{\sqrt{i}}{\varkappa_2}+\sqrt{k}+\frac{k}{\sqrt{i\varkappa_2}})\epsilon+\\
    &\hspace{3cm}\frac{k}{\sqrt{i\varkappa_2}}+\sqrt{i}+k\delta\Big)\\
    &\hspace{5cm}\leq 2(\varkappa_1+1)e^{-2\epsilon^2}
\end{align*}
As such, we have $\mathds{E}[\frac{|j\hat{c}(q)-c_{\mD{j}}(q)|-(\frac{k}{\sqrt{i\varkappa_2}}+\sqrt{i}+k\delta)}{\sqrt{i}+\frac{\sqrt{i}}{\varkappa_2}+\sqrt{k}+\frac{k}{\sqrt{i\varkappa_2}}}]\leq \varkappa_3$, for some universal constant $\varkappa_3$ so that, 
$\mathds{E}[|j\hat{c}(q)-c_{\mD{j}}(q)|]\leq\frac{k}{\sqrt{i\varkappa_2}}+\sqrt{i}+k\delta+\varkappa_3\sqrt{i}+\frac{\varkappa_3\sqrt{i}}{\varkappa_2}+\varkappa_3\sqrt{k}+\frac{\varkappa_3k}{\sqrt{i\varkappa_2}}$. Assuming $k\leq i$, we have $$\mathds{E}[|j\hat{c}(q)-c_{\mD{j}}(q)|]\leq\varkappa(\sqrt{i}+\sqrt{k}+k\delta),$$ For some universal constant $\varkappa$.

Now if $\delta\geq \frac{2\varkappa}{\sqrt{i}}$ we let $k=\frac{\phi-\varkappa}{2\varkappa\delta}\sqrt{i}$ and otherwise set $k=\rho i$ for $\rho=\min\{(\frac{\phi-\varkappa}{\varkappa(1+2\varkappa)})^2, 1\}$. 

First, consider the case where $\delta\geq \frac{1}{\sqrt{j}}$. Consider the error $\epsilon=\phi\sqrt{j}\geq \phi\sqrt{i}$, so it suffices to show that the error is at most $\phi\sqrt{i}$. Let $k=\frac{\phi-\varkappa}{2\varkappa\delta}\sqrt{i}$. Substituting this in, we want to show 
$\sqrt{\frac{\varkappa(\phi-\varkappa)}{2\delta}\sqrt{i}}-\frac{\phi-\varkappa}{2}\sqrt{i}\leq 0$. Indeed, for $\delta\geq \frac{2\varkappa}{\sqrt{i}}$, we have $\frac{\varkappa(\phi-\varkappa)}{2\delta}\sqrt{i}\leq\frac{(\phi-\varkappa)}{4}i$, so that 
\begin{align*}
    \sqrt{\frac{\varkappa(\phi-\varkappa)}{2\delta}\sqrt{i}}-\frac{\phi-\varkappa}{2}\sqrt{i}&\leq \frac{1}{2}\sqrt{(\phi-\varkappa)i}-\frac{\phi-\varkappa}{2}\sqrt{i}\\
    =&\frac{1}{2}\sqrt{(\phi-\varkappa)i}(1-\sqrt{\phi-\varkappa})\\
    \leq&0
\end{align*}
Which proves $\mathds{E}[|j\hat{c}(q)-c_{\mD{j}}(q)|]\leq\phi\sqrt{j}$ whenever $\delta\geq\frac{2\varkappa}{\sqrt{i}}$ and $\phi-\varkappa\geq 1$.

If $\delta\leq \frac{2\varkappa}{\sqrt{i}}$, we set $k=\rho i$ for $\rho=\min\{(\frac{\phi-\varkappa}{\varkappa(1+2\varkappa)})^2, 1\}$. We have
\begin{align*}
    \varkappa(\sqrt{i}+\sqrt{\rho i}+\rho i\delta)\leq \varkappa\sqrt{i}(1+\sqrt{\rho}+2\sqrt{\rho}\varkappa)
\end{align*}
So that we need to ensure $1+\sqrt{\rho}(1+2\varkappa)\leq\frac{\phi}{\varkappa}.$ Observe that $\sqrt{\rho}\leq\frac{\phi-\varkappa}{\varkappa(1+2\varkappa)}$ implies the above, so that setting $\rho$ as above proves the result in this case.

\subsubsection{One Dimension (Theorem~\ref{thm:dynamic_ce_1d_oracle})}
\textbf{Construction}. We build a B-tree like data structure. However, in addition to the content of each node, each node also keeps a counter of the number of elements inserted into the node. After each insertion, if a leaf node has more than $k$ elements, the node is split into two, for $k=\frac{\epsilon^2}{4(\varkappa+1)^2}$. Thus, leaf nodes cover between $\frac{k}{2}$ to $k$ elements, while the rest of the tree has its own fanout $B$. Leaf nodes do not store the elements associated with them, but build models to answer queries. To answer a query, the tree is traversed similar to typical range query answering with a B-tree. However, if a node is fully covered in a range, then the number of insertions in the node is used to answer queries. Otherwise, the node is recursively searched until reaching a leaf node. We will have at most $2$ leaf nodes reached that will be partially covered by the query range. Finally, the model of each node are constructed by using the construction in Theorem~\ref{thm:dynamic_ce_hd_oracle} with $\phi=\varkappa+1$. 

\textbf{Query Time and Space Consumption}. Each query will take $O(\log n+2\mathcal{T}_k^\mathfrak{X})$ where $\log n$ is due to the tree traversal and $2\mathcal{T}_k^\mathfrak{X}$ for the two model inferences needed. Moreover, the total space consumption is $O(\frac{n}{k}\mathcal{S}_k^\mathfrak{X}+\frac{n}{k}\log n)$


\textbf{Insertion Complexity}. Each leaf node will start with $\frac{k}{2}$ elements and will be split whenever it reaches $k$ elements. Insertion of the $\frac{k}{2}$ elements in a node cost $O(k+\max\{\delta\sqrt{k}, 1\}\mathcal{B}_{k}^\mathfrak{X})$. Given $n$ insertions, we have $\frac{2n}{k}$ insertions of $\frac{k}{2}$ elements in the nodes, so that the amortized cost of rebuilds is $O(\frac{1}{n}\frac{n}{k}(k+\max\{\delta\sqrt{k}, 1\}\mathcal{B}_{k}^\mathfrak{X}))=O(\max\{\frac{\delta}{\sqrt{k}}, \frac{1}{k}\}\mathcal{B}_{k}^\mathfrak{X}))$. Furthermore, traversing the tree nodes costs $O(\log n)$ per insertion, so that amortized insertion cost is $O(\max\{\frac{\delta}{\sqrt{k}}, \frac{1}{k}\}\mathcal{B}_{k}^\mathfrak{X})+\log n)=O(\max\{\frac{\delta}{\epsilon}, \frac{1}{\epsilon^2}\}\mathcal{B}_{\epsilon^2}^\mathfrak{X})+\log n)$.



\textbf{Accuracy}. Setting $\phi=\varkappa+1$ in Theorem~\ref{thm:dynamic_ce_hd_oracle} and having $k\leq\frac{\epsilon^2}{4(1+\varkappa)^2}$ ensures that the expected error of the each model is at most $\phi\sqrt{k}=\frac{\epsilon}{2}$. Since each query is answered by making two model calls, the total expected error for answering queries is at most $\epsilon$ as required.

\begin{algorithm}[t]
\begin{algorithmic}[1]
\Require Query $\vq$, dimension to refine, $i$, set of models, $M$, and set of partition points $S$ 
\Ensure Estimate to cardinatliy of $\vq$
\Procedure{Query}{$\vq$, $i$, $S$, $M$}
    \If{$i=0$}
        \State\Return use grid to answer $\vq$
    \EndIf
    \State $i_l\leftarrow$ index of $\vq[i][0]$ in $S[i]$
    \State $i_u\leftarrow$ index of $\vq[i][1]$ in $S[i]$
    \If{$i_u=i_l$}
        \State\Return $M[i][i_l](\vq)$
    \EndIf 
    \State $\vq_u\leftarrow \vq$
    \State $\vq_u[i][0]\leftarrow S[i_u]$
    \State $\vq_l\leftarrow \vq$
    \State $\vq_l[i][1]\leftarrow S[i_l+1]$
    \If{$i_u=i_l+1$}
        \State\Return $M[i][i_l](\vq_l)+M[i][i_u](\vq_u)$
    \EndIf 
    \State $\vq[i][0]\leftarrow S[i_l+1]$
    \State $\vq[i][1]\leftarrow S[i_u]$
    \State\Return \textsc{Query}($\vq$, $i-1$, $S$, $M$)+$M[i][i_l](\vq_l)+M[i][i_u](\vq_u)$
\EndProcedure
\caption{Cardinality Estimation with Grid}\label{alg:hd_ce_any_error}
\end{algorithmic}
\end{algorithm}

\subsubsection{High dimension with any accuracy}\label{appx:hd_ce_oracle_anyaccuracy}
Here we also discuss how we can use models to answer queries to arbitrary accuracy in high dimensions. We note that, as we see here, building data structures to answer queries is high dimension is difficult. We discuss this result only in the static setting. 

\begin{lemma}\label{lemma:hd_ce_oracle_anyaccuracy}
There exists a learned model that can answer cardinality estimation query with error up to $\epsilon$ with query time $O(\mathcal{T}_{(\frac{\epsilon}{2d})^2}^{\mathfrak{X}}+(\frac{4d^2}{\epsilon^2})^d\Pi_ik_i)$ and space complexity $O(\frac{4d^2n}{\epsilon^2}\mathcal{S}_{(\frac{\epsilon}{2d})^2}^{\mathfrak{X}}+(\frac{4d^2n}{\epsilon^2})^d)$, where $k_i$ is the cardinality of the query in the $i$-th dimension.
\end{lemma}
Cardinality of a query in the $i$-th dimension is the number of points the would query we only consider the $i$-th dimension. 

\textbf{Construction}. Assume we would like to obtain accuracy $\epsilon$. We build a grid and materialize the exact result in each cell. Then, for queries, where part of a query partially overlaps a cell, we also build models to answer queries. Thus, a query is decomposed into parts that fully contain cell and parts that don't which are answered by models. 

Split the $i$-th dimension into $k=\frac{4d^2n}{\epsilon^2}$ partitions, with each partition containing $(\frac{\epsilon}{2d})^2$ points. Let $S[i]=\{s_1^i, ..., s_k^i\}$ be the partition points in the $i$-th dimension, that is, for all $j$ we have for $P[i, j]=\{\vp\in \mD, s_j^i\leq \vp_i< s_{j+1}^i\}$, $|P[i, j]|=(\frac{\epsilon}{2d})^2$. Using theorem~\ref{thm:dynamic_ce_hd_oracle} to build a model for each set of points in $P$, we have that the expected error of each is $O(\frac{\epsilon}{2d})$. The models are stored in $M$, with $M[i][j]$ denoting the model corresponding to $j$-th partition in the $i$-th dimension. Now, to answer a query, we first decompose it into $2d+1$ queries. $2d$ of the queries are answered by models, which reduce the original query to one that matches all facets of the grid cells. Then, the grid cells are used to answer the final query, and the answer is combined with the model estimates to find final query answer estimate. 

This is presented in Alg.~\ref{alg:hd_ce_any_error}. The decomposition of the query is done by recursively moving the upper and lower facets of the query hyperretangle in the $d$-th dimensions to aligh with the grid cell. Thus, in the $d$-th dimensions, if the closest grid partition points, respectively larger and smaller than $q[d][0]$ and $q[d][1]$ (the lower and upper bound of the query in $d$-th dimension) are $s_i$ and $s_j$, we decompose the query into three queries: $q_1$, $q_2$ and $q_3$, all the same as $q$ but $q_1[d][1]=s_i$, $q_2[d][0]=s_j$ and $q_3[d][0]=s_i$, $q_3[d][1]=s_j$. Then, learned models are used to answer $q_1$ and $q_2$, while $q_3$ is further recursively decomposed along its $d-1$-th dimension (and after full decomposition is answered using the grid). Note that $q_1$ and $q_2$ can now be answered using models, because by grid construction, they fall in a part of the space with at most $(\frac{\epsilon}{2d})^2$ points. 

\textbf{Accuracy}. Grid cells are exact, and, as discussed above each model is built on a dataset of size at  most $(\frac{\epsilon}{2d})^2$, so that it will have expected error $O(\frac{\epsilon}{2d})$. Thus, combining the error of the $2d$ queries, the total model error is $\epsilon$ as desired. 

\textbf{Query Time and Space Complexity}. There are $2d$ model calls, each model call costing $\mathcal{T}_{(\frac{\epsilon}{2d})^2}^{\mathfrak{X}}$. Furthermore, if the $i$-th dimension of the query covers $k_i$ points, then total of at most $\frac{4d^2 k_i}{\epsilon^2}$ partitions in the $i$-th dimension intersect the query, so that the total number of cells traversed will be $(\frac{4d^2}{\epsilon^2})^d\Pi_ik_i$. Thus, total query time is $O(\mathcal{T}_{(\frac{\epsilon}{2d})^2}^{\mathfrak{X}}+(\frac{4d^2}{\epsilon^2})^d\Pi_ik_i)$. Furthermore, the total cost of storing the models is $\frac{4d^2n}{\epsilon^2}\mathcal{S}_{(\frac{\epsilon}{2d})^2}^{\mathfrak{X}}$, and the cost of the grid is $(\frac{4d^2n}{\epsilon^2})^d$. Thus, total space complexity is $O(\frac{4d^2n}{\epsilon^2}\mathcal{S}_{(\frac{\epsilon}{2d})^2}^{\mathfrak{X}}+(\frac{4d^2n}{\epsilon^2})^d)$.

%% file: proof_sorting.tex
\subsubsection{Using Learned Model (Theorem~\ref{thm:sorting_oracle})}

\begin{algorithm}[t]
\begin{algorithmic}[1]
\Require An array $A$ of length $n$ to be sorted
\Ensure The sorted array
\Procedure{Sort}{$A$}
    \If{$n\leq \varkappa$}
        \State \Return $\textsc{MergeSort(A)}$
    \EndIf
    \State $S\leftarrow$ random sample of $A$ of size $\sqrt{n}$
    \State $S\leftarrow \textsc{MergeSort}(S)$
    \State $k\leftarrow n^{\frac{1}{8}}$
    \State $\hat{f}\leftarrow\mathcal{A}(S)$
    \State $B\leftarrow$ array of size $k$
    \State $B_{min}, B_{max}\leftarrow$ arrays tracking min/max $B[i]\;\forall i$
    \For{$i$ in $n$}
        \State $B[\floorx{\frac{k\hat{f}(A[i])}{n}}].append(A[i])$
        \State Update $B_{min}$, $B_{max}$ for bucket $\floorx{\frac{k\hat{f}(A[i])}{n}}$
    \EndFor
    \For{$i$ \textbf{in} $k-2$}
        \If {$B_{max}[i]>B_{min}[i+(2\varkappa+1)]$}
            \State \Return $\textsc{MergeSort}(A)$
        \EndIf
    \EndFor
    \For{$i$ \textbf{in} $k$}
        \If{$|B[i]|\geq (2\varkappa+1)n^{\frac{4}{5}}$}
            \State \Return $\textsc{MergeSort}(A)$
        \Else
            \State $B[i]\leftarrow \textsc{Sort}(B[i])$
        \EndIf
    \EndFor
    \State \Return $\textsc{Merge}(B)$ \Comment{Alg.~\ref{alg:sort_merge}}
\EndProcedure
\caption{Learned Sorting}\label{alg:dynamic_learned_index}
\end{algorithmic}
\end{algorithm}

\begin{algorithm}[t]
\begin{algorithmic}[1]
\Require An array of sorted buckets
\Ensure Buckets merged into a sorted array
\Procedure{Merge}{$B$}
    \State $A_s\leftarrow$ empty array of size $n$
    \State $A_s[:len(B[1])]\leftarrow B[1]$
    \State $j\leftarrow len(B[1])$
    \For{$b\leftarrow 2$ to $k$}\Comment{Merges $A_s[:j]$ with $B[b]$}
        \State $j\leftarrow j+len(B[b])$\Comment{Iterator for $A_s$}
        \State $i\leftarrow len(B[b])$\Comment{Iterator for $B[b]$}
        \While {$i>0$}
            \If {$B[b][i]>A_s[j]$}
                \State $A_s[j+i]\leftarrow B[b][i]$
                \State $i${-}{-}
            \Else 
                \State $A_s[j+i]\leftarrow A_s[j]$
                \State $j${-}{-}
            \EndIf
        \EndWhile
        \State $j\leftarrow j+len(B[b])$
    \EndFor    
    \State \Return $A_s$
\EndProcedure
\caption{Merge Step}\label{alg:sort_merge}
\end{algorithmic}
\end{algorithm}

\textbf{Algorithm.} The algorithm is presented in Alg.~\ref{alg:dynamic_learned_index}. A sample of the array is first created and a model is built using the sample. Then, using the model, the array is split into $n^{\frac{1}{8}}$ buckets, where we theoretically show, using such a number of buckets, based on the accuracy of the model and with high probability, merging the buckets can be done by in linear time (because there will be limited overlap between the buckets) and each bucket will not be too big. Indeed, we first make sure the two properties mentioned before hold (otherwise the algorithm quits and reverts to merge sort), and then proceed to merge the buckets. 

\textbf{Correctness}. If all the created buckets are sorted, the merge step simply merges them and thus returns a sorted array correctly. At the base case, merge sort is used, so the buckets will be sorted correctly. Thus, using the invariant above, the algorithm is correct. 

\textbf{Time Complexity}.
Consider sorting an array $A\sim\chi$ of size $n$. We take a subset $S$, of size $\sqrt{n}$ from $A$ without checking the elements to preserve the i.i.d assumption.  We sort them and use the algorithm $\mathcal{A}$ to obtain a model $\hat{r}$. We have that \begin{align}
    \mathds{P}(\normx{\hat{r}-r_{\chi}}_{\infty}\geq \sqrt{n}\epsilon_1)\leq \varkappa_1e^{-\varkappa_2(\epsilon_1n^{-\frac{1}{4}}-1)^2.}
\end{align}
Note that
\begin{align*}
\normx{\hat{r}-r_A}_\infty&\leq \normx{\hat{r}-r_{\chi}}_\infty+\normx{r_{\chi}-r_A},
\end{align*}
And, by DWK,
$$\mathds{P}(\normx{r_A-r_\chi}_\infty\geq \epsilon_3)\leq 2e^{-2(\frac{\epsilon_3}{\sqrt{n}})^2}.$$
\if
so that 
\begin{align*}
\mathds{P}(\normx{\hat{r}-r_{\chi}}_{\infty}\geq \epsilon_1 \text{\;or\;} \normx{r_A-r_\chi}\geq \epsilon_3)\\\leq \varkappa_1 e^{-\varkappa_2\sqrt{n}\epsilon_1^2}+\varkappa_1e^{-\varkappa_2n\epsilon_3^2}
\end{align*}
\fi
Set $\epsilon_1=n^{\frac{1}{4}}(\sqrt{\frac{\log\log n}{\varkappa_2}}+1)$ and $\epsilon_3=\sqrt{\frac{n\log\log n}{2}}$, we have
\begin{align*}
\mathds{P}(\normx{\hat{r}-r_{\chi}}_{\infty}\geq \sqrt{n}\epsilon_1 \text{\;or\;} \normx{r_A-r_\chi}\geq \epsilon_3)\\
\leq \varkappa_1e^{-\log\log n}+2e^{-\log\log n}\\
=\frac{\varkappa_1+2}{\log n}
\end{align*}
Thus, 
$$
\mathds{P}(\normx{\hat{r}-r_A}\geq \sqrt{n}\epsilon_1+\epsilon_3)\leq \frac{\varkappa_1+2}{\log n}
$$
and thus, whenever $\log\log n\geq 2$ and for $\varkappa=\frac{1}{\sqrt{2}}+\frac{1}{\sqrt{\varkappa_2}}$
$$
\mathds{P}(\normx{\hat{r}-r_A}_\infty\geq \varkappa n^{\frac{3}{4}}\sqrt{\log\log n})\leq \frac{\varkappa_1+2}{\log n}.
$$
To simplify, observe that $n^{3/4}\sqrt{\log\log n}\leq n^{4/5}$ for $n\geq e$ so that 
$$
\mathds{P}(\normx{\hat{r}-r_A}_\infty\geq \varkappa n^{\frac{4}{5}})\leq \frac{\varkappa_1+2}{\log n}.
$$

Recall that we use $k$ buckets and consider the $i$-th bucket. The elements, $x$, assigned to it must have $\frac{i}{k}\leq \frac{1}{n}\hat{r}(x)<\frac{i+1}{k}$. Combining this with the above, we have that whenever $\normx{\hat{r}-r_A}_\infty\leq \varkappa n^{\frac{4}{5}}$ holds, for the elements $x$ in the $i$-th bucket, we must have $\frac{1}{n}\hat{r}(x)\geq \frac{i}{k}-\varkappa n^{-\frac{1}{5}}$ and that $r_A(x)\leq \frac{i+1}{k}+\varkappa n^{-\frac{1}{5}}$. Therefore, we must have $r_A(x)\in [\frac{in}{k}-\varkappa n^{\frac{4}{5}}, \frac{(i+1)n}{k}+\varkappa n^{\frac{4}{5}}]$. There are at most $\frac{n}{k}+2\varkappa n^{\frac{4}{5}}$ elements in this set. Setting $k=n^{\frac{1}{5}}$, we have that whenever $\normx{\hat{r}-r_A}_\infty\leq \varkappa n^{\frac{4}{5}}$ holds, all buckets will have at most $(2\varkappa+1)n^{\frac{4}{5}}$ elements. Thus, the probability that a bucket will have more than $(2\varkappa+1)n^{\frac{4}{5}}$ elements is at most $\frac{\varkappa_1+2}{\log n}$.

Furthermore, the largest element in the $i$-th bucket or before will have $r_A(x)< (i+1+\varkappa)n^{\frac{4}{5}}$ and the smallest element in the $i+2\varkappa+1$-th bucket or after will have $r_A(x)\geq (i+1+\varkappa)n^{\frac{4}{5}}$, so that the content of buckets up to $i$ are less than the content of the buckets from $i+2\varkappa+1$ onwards. Thus, whenever $\normx{\hat{r}-r_A}_\infty\leq \varkappa n^{\frac{4}{5}}$ holds, if all the buckets are sorted, then the algorithm takes at most $\sum_{i=1}^k\sum_{j=0}^{2\varkappa+1}|B_{i-j}|$ number of operations to merge the sorted array, which is $O(n)$, where $|B_i|$ is the number of elements in the $i$-th bucket. 

\if 
Now use $n\hat{f}$ to group the elements into $k$ groups, where an element is assigned to the group $\floorx{k\hat{f}}$. Observe that for an element to be assigned to the $i$-th group, we need to have

insert the elements into the array, and group every $3n^{\frac{3}{4}}\sqrt{\log\log n}$ elements together. Let $k = 3n^{\frac{3}{4}}\sqrt{\log\log n}$,  Let $\bar{N}_{k, i}$ be the size of the number of points in the $i$-th group. Also consider grouping the slots of every $3$ consecutive groups, so that $N_{3k, i}=\sum_{j=0}^{3k}N_{i+j}$. Note that $\bar{N}_{k, i}\leq N_{3k, i-1} $. 
\fi

Finally, let $T_n$ be the expected number of operations it takes to sort an array with $n$ elements i.i.d sampled from some a distribution learnable class. Recall that we do merge sort if $\normx{\hat{r}-r_A}_\infty\leq \varkappa n^{\frac{4}{5}}$ does not hold, and recursively sort the array if it does. Thus, whenever we recursively sort an array, the array will have at most $(2\varkappa+1)n^{\frac{4}{5}}$ i.i.d elements, distributed from a conditional distribution of the original distribution. Thus, we have
\begin{align*}
    T(n)&\leq O(\mathcal{B}_{\sqrt{n}}^\mathfrak{X}+\mathcal{T}_{\sqrt{n}}^\mathfrak{X}n)+\mathds{P}(\text{merge sort})n\log n+\\&\qquad\qquad\qquad\mathds{P}(\text{recursively sort})n^{\frac{1}{5}}T((2\varkappa+1)n^{\frac{4}{5}})\\
    &\leq O(\mathcal{B}_{\sqrt{n}}^\mathfrak{X}+\mathcal{T}_{\sqrt{n}}^\mathfrak{X}n)+\frac{\varkappa+2}{\log n}n\log n+\\&\hspace{4cm} n^{\frac{1}{5}}T((2\varkappa+1)n^{\frac{4}{5}})\\
    &=O(\mathcal{T}_{\sqrt{n}}^\mathfrak{X}n\log\log n+\\&\hspace{2cm}\sqrt{n}\mathcal{B}_{\sqrt{n}}^\mathfrak{X}+\sum_{i=0}^{\log\log n}n^{1\shortminus\frac{1}{2}(\frac{4}{5})^{i}}\mathcal{B}_{n^{\frac{1}{2}(\frac{4}{5})^{i}}}^\mathfrak{X}).
\end{align*}
\textbf{Space Comlexity}. First, observe that we only create one model at a time, so the maximum size used for modeling is $\mathcal{S}_{\sqrt{n}}^\mathfrak{X}+\sqrt{n}\log n$. Moreover, the depth of recursion is at most $O(\log\log n)$, and the overhead of storing $B$ is dominated by the first of recursion, whose overhead is $O(n^{\frac{1}{8}}\log n+O(n\log n))$, giving overall space overhead of $O(\mathcal{S}_{\sqrt{n}}^\mathfrak{X}+n\log n)$

\subsubsection{Using data distribution (Theorem~\ref{thm:sorting_oracle_accurate})}

\begin{algorithm}[t]
\begin{algorithmic}[1]
\Require An unsorted array $A$ of size $n$
\Ensure A sorted array
\Procedure{Sort}{$A$}
    \If{$n\leq 10$}
        \State\Return{$\textsc{MergeSort}(A)$}
    \EndIf
    \State $A'\leftarrow$ new array, $A'[i]$ initialized as linked list, $\forall i$
    \For{$i\leftarrow 1$ \textbf{to} $n$}
        \State $i'\leftarrow \ceilx{\hat{r}(A[i])}$
        \State $A'[i'].append(A[i])$
    \EndFor
    \State \Return $\textsc{Merge}(B)$ \Comment{Alg.~\ref{alg:sort_merge}}
\EndProcedure
\caption{Sorting Using Distribution Model}\label{alg:dist_sort}
\end{algorithmic}
\end{algorithm}

Alg.~\ref{alg:dist_sort} shows how to sort an array given an approximate model of the data distribution $\hat[r]$. The algorithm is very similar to Alg.~\ref{alg:dist_sort}, but uses $n$ buckets and merges each bucket using merge sort (and thus no recursive sorting of the buckets).

\textbf{Correctness}. The algorithm creates buckets, sorts them independently. The sort is done by merge sort so it is correct, and thus merging the sorted buckets creates a sorted array. 

\textbf{Running time}
Let ${T}(A)$ be the number of operations the algorithm performs on an array $A$, and let $\mathcal{T}(n)=\mathds{E}_{A\sim\chi^n}[{T}(A)]$ be the expected run time of the algorithm on an input of size $n$. 

First, assume we use $\floorx{nr_\chi(x)}$ to map an element $x$ to a location in the array $S$. After the mapping, we study the expected time to sort the elements in $S[i:j]$ where $j=i+k$ and $k>0$. Note that the probability that an element is mapped to location $[i:j]$, i.e., $\mathds{P}_{x\sim \chi}(\frac{i}{n}\leq r_\chi(x)<\frac{j}{n})$, is $\frac{k}{n}$. Let $N_{i:j}=|S[i:j]|$ be the number of elements mapped to $S[i:j]$. We have that 
$$\mathds{P}_{A\sim\chi^n}(N_{i:j}=z)=\mathds{C}(n, z)(\frac{k}{n})^z(1-\frac{z}{n})^{n-z}.$$ 
Thus, we have
\begin{align*}
\mathds{E}_{A
\sim\chi^n}[N_{i:j}\log(N_{i:j})]&=\sum_{z=1}^{n}\mathds{P}_{A\sim\chi^n}(N_{i:j}=z)[z\log(z)]\\
&=\sum_{z=1}^{n}\mathds{C}(n, z)(\frac{k}{n})^z(1-\frac{k}{n})^{n-z}z\log z\\
&=\sum_{i=1}^{4ek}\mathds{C}(n, i)(\frac{k}{n})^i(1-\frac{k}{n})^{n-i}i\log i\\&\;+\sum_{i=4ek}^{n}\mathds{C}(n, i)(\frac{k}{n})^i(1-\frac{k}{n})^{n-i}i\log i
\end{align*}
For the first part of the summation, we have
\begin{align*}
\sum_{i=1}^{4ek}\mathds{C}(n, i)&(\frac{k}{n})^i(1-\frac{k}{n})^{n-i}i\log i \\ &\leq\log(4ek)\sum_{i=1}^{4ek}\mathds{C}(n, i)(\frac{k}{n})^i(1-\frac{k}{n})^{n-i}i\\
&\leq k\log(4ek).
\end{align*}
For the second part, we have
\begin{align*}
    &\sum_{i=4ek}^{n}\mathds{C}(n, i)(\frac{k}{n})^i(1-\frac{k}{n})^{n-i}i\log i\\
&\leq\sum_{i=4ek}^{n}\frac{1}{\sqrt{i}}(\frac{en}{i})^i(\frac{k}{n})^i(1-\frac{k}{n})^{n-i}i\log i\\
&= \sum_{i=4ek}^{n}(\frac{ekn}{i(n-k)})^i(1-\frac{k}{n})^{n}\sqrt{i}\log i\\
&\leq \sum_{i=4ek}^{n}(\frac{2ek}{i})^i(1-\frac{k}{n})^{n}\sqrt{i}\log i\\
&\leq \sum_{i=4ek}^{n}(\frac{1}{2})^ii\\
&\leq 2.\\
\end{align*}
So that 
$$
\mathds{E}_{A
\sim\chi^n}[N_{i:j}\log(N_{i:j})]\leq (j-i)\log (4e(j-i))+2.
$$
Now recall that we use $\floorx{\hat{r}}$ with error $\normx{\hat{r}-nr_\chi}_\infty\leq \epsilon$ to map the elements to an array $S'$. 
Thus, if, for any $x$,  $\floorx{nr_\chi(x)}=j$, $\hat{r}\in \{j-\floorx{\epsilon}, ..., j+\ceilx{\epsilon}\}$. 
Let $\bar{N}_{i:j}$ be the number of elements mapped to positions $[i:j]$ using $\hat{r}$. Note that $\bar{N}_{i:j}\leq N_{i-\epsilon:j+\epsilon}$. Thus, dividing $S'$ into groups of $\epsilon$ and sorting each separately, we have that the total cost of sorting the groups is
\begin{align*}
    \sum_{j=0}^{\frac{n}{\epsilon}}{\mathds{E}_{A\sim\chi^n}[\bar{N}_{j\epsilon: (j+1)\epsilon}\log(\bar{N}_{j\epsilon:(j+1)\epsilon})]}\\\leq \sum_{j=0}^{\frac{n}{\epsilon}}{\mathds{E}_{A\sim\chi^n}[{{N}_{(j-1)\epsilon:(j+2)\epsilon}}\log({N}_{(j-1)\epsilon:(j+2)\epsilon})]}\\
    \leq O(\epsilon\log \epsilon)
\end{align*}
Finally, note that $r_\chi$ is a non-decreasing function. Therefore, if $x\leq y$, we have $\floorx{nr_\chi(x)}\leq \floorx{nr_\chi(y)}$. Given that $\normx{\hat{r}-nr_\chi}_\infty\leq \epsilon$, we have that if $x\leq y$, $\hat{r}(y)\geq \hat{r}(x)-2\epsilon$. Consequently, if $x$ is mapped to the $i$-th group by $\hat{r}$, all elements in the $j$-th group with $j<i-2$ are less than $x$. This means, to merge the sorted groups, we start with the first group and iteratively merge the next group with the merged array so far. Performing each merge from the end of the two sorted arrays (as done in merging using learned data distribution), each merge will cost at most $3\epsilon$, so the total cost of merging all the $\frac{n}{\epsilon}$ sorted groups is $O(\frac{n}{\epsilon}\epsilon)=O(n)$. Putting everything together,r when each model call costs $\mathcal{T}_n^\mathfrak{X}$, we have that expected time complexity of the algorithm is $O(n\mathcal{T}_n^\mathfrak{X}+n\log\epsilon)$. Moverover the space overhead of the algorithm is $O(n\log n+\mathcal{S}_n^\mathfrak{X})$ where the $\log n$ factor is to keep a pointer to the elements of the array (instead of copying them).